\documentclass[nohyperref]{article}

\usepackage{microtype}
\usepackage{graphicx}
\usepackage{subfigure}
\usepackage{booktabs} 

\usepackage{hyperref}

\newcommand{\myparagraph}[1]{\textbf{#1}}

\usepackage[accepted]{icml2022}

\usepackage{amsmath}
\usepackage{amssymb}
\usepackage{mathtools}
\usepackage{amsthm}
\usepackage{lipsum}
\usepackage{enumitem}
\usepackage{mathtools}
\usepackage{cuted}

\usepackage[capitalize,noabbrev]{cleveref}

\theoremstyle{plain}

\theoremstyle{definition}

\theoremstyle{remark}

\usepackage[textsize=tiny]{todonotes}

\icmltitlerunning{Cycle Representation Learning for Inductive Relation Prediction}

\begin{document}

\twocolumn[
\icmltitle{Cycle Representation Learning for Inductive Relation Prediction}



\icmlsetsymbol{equal}{*}

\begin{icmlauthorlist}
\icmlauthor{Zuoyu Yan}{pku}
\icmlauthor{Tengfei Ma}{ibm}
\icmlauthor{Liangcai Gao}{pku}
\icmlauthor{Zhi Tang}{pku}
\icmlauthor{Chao Chen}{sbu}
\end{icmlauthorlist}

\icmlaffiliation{pku}{Wangxuan Institute of Computer Technology, Peking University, Beijing, China}
\icmlaffiliation{sbu}{Department of Biomedical Informatics, Stony Brook University, New York, USA}
\icmlaffiliation{ibm}{T. J. Watson Research Center, IBM, New York, USA}

\icmlcorrespondingauthor{Chao Chen}{chao.chen.1@stonybrook.edu}
\icmlcorrespondingauthor{Liangcai Gao}{glc@pku.edu.cn}
\icmlcorrespondingauthor{Zhi Tang}{tangzhi@pku.edu.cn}

\icmlkeywords{Inductive Relation Prediction, Topological Data Analysis, Cycle Basis, Homology}

\vskip 0.3in
]



\printAffiliationsAndNotice{}  

\begin{abstract}
In recent years, algebraic topology and its modern development, the theory of persistent homology, has shown great potential in graph representation learning. In this paper, based on the mathematics of algebraic topology, we propose a novel solution for inductive relation prediction, an important learning task for knowledge graph completion. To predict the relation between two entities, one can use the existence of rules, namely a sequence of relations. Previous works view rules as paths and primarily focus on the searching of paths between entities. The space of rules is huge, and one has to sacrifice either efficiency or accuracy. In this paper, we consider rules as cycles and show that the space of cycles has a unique structure based on the mathematics of algebraic topology. By exploring the linear structure of the cycle space, we can improve the searching efficiency of rules. We propose to collect cycle bases that span the space of cycles. We build a novel GNN framework on the collected cycles to learn the representations of cycles, and to predict the existence/non-existence of a relation. Our method achieves state-of-the-art performance on benchmarks.
\end{abstract}

\section{Introduction}
\label{sec:intro}

Knowledge graphs (KGs) are graph-structured knowledge bases that integrate human knowledge through relational triplets. In a KG, nodes represent entities and edges represent relational triplets connecting them. A relational triplet is defined as $(e_h, r, e_t)$, where $e_h$ and $e_t$ are the head and tail entities respectively, and $r$ is the relation between them. KGs have been used in many problems such as recommendation systems~\citep{wang2018ripplenet}, question answering~\citep{huang2019knowledge, zhang2018variational}, biomedical research~\citep{zhao2020biomedical,zhu2020knowledge}, and zero-shot learning~\citep{kampffmeyer2019rethinking}.

Due to the limitation of human knowledge and data extraction algorithms, we cannot thoroughly define and excavate all the entities and relations in a KG~\citep{chen2020knowledge}. The incomplete structures and contents of KGs can significantly benefit from an automatic completion algorithm. Early works~\citep{bordes2013translating, yang2014embedding, sun2019re} focus on incorporating the attributes of entities. Recent works develop models that are agnostic of entity attributes. They can handle new entities and dynamic KGs, which are quite common.

\begin{figure*}[hbtp]
	\centering
	\includegraphics[width=\textwidth]{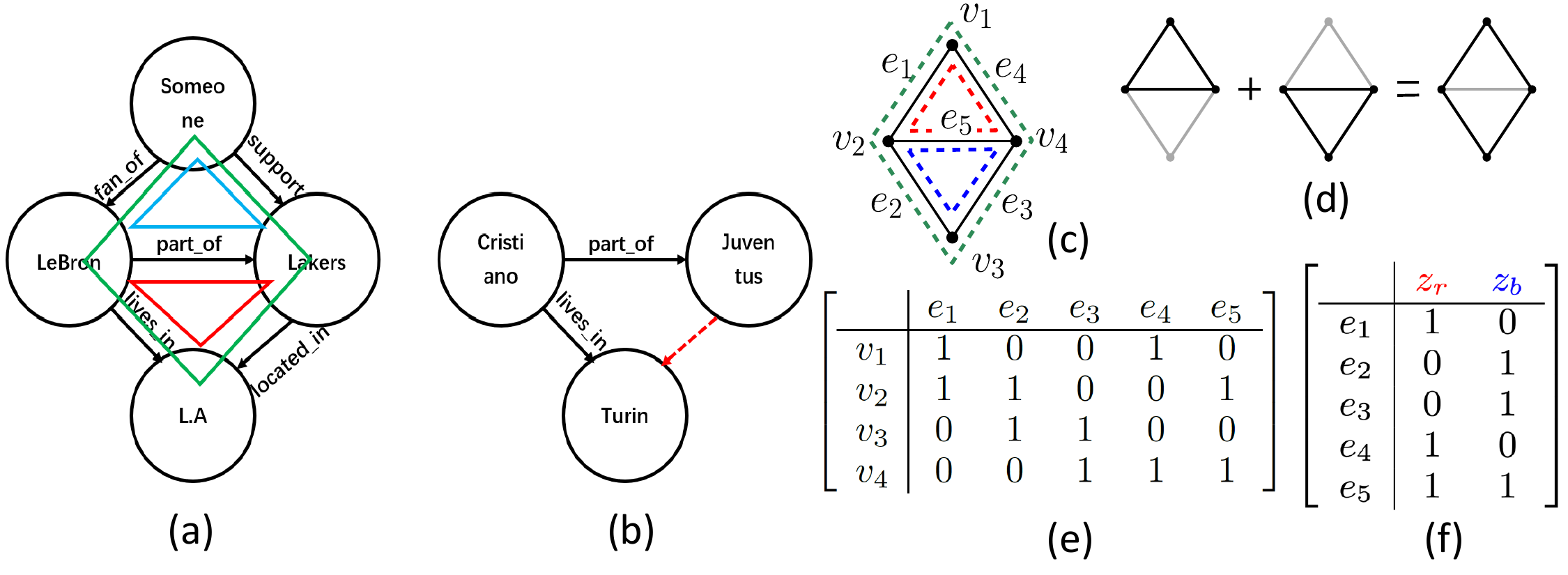}
	\vspace{-.3in}
	\caption{\textbf{(a)} and \textbf{(b):} examples of inductive relation prediction. \textbf{(c)-(f):} illustrations for cycle space and cycle basis. \textbf{(c):} A sample graph with three nontrivial cycles, $z_r$, $z_b$ and $z_g$ (highlighted with red, blue and green colors). Any two of the three will form a cycle basis. \textbf{(d):} Mod-2 addition of cycles. \textbf{(e):} the boundary matrix $\partial$ of the graph in (c). Any cycle $z$ satisfies $\partial z=0$. \textbf{(f):} the cycle incidence matrix of the graph in (c). We show all three columns corresponding to all three nontrivial cycles ($z_r$, $z_b$ and $z_g$). In our algorithm, we only pick the columns corresponding to a chosen cycle basis, e.g., the first two columns when the chosen basis is $\{z_r,z_b\}$.}
	\label{fig:motiv}
	\vspace{-.15in}
\end{figure*}

These entity-agnostic methods~\citep{yang2017differentiable, sadeghian2019drum, teru2020inductive} are called \emph{inductive relation prediction} methods.  
They predict missing triplets by learning logical rules in KGs. For example, from the KG shown in Figure~\ref{fig:motiv}(a), we can learn the rule: 
\begin{multline}
\label{equa: ind}
\exists X, (X, \text{ \textit{part\_of} }, Y) \wedge (X, \text{ \textit{lives\_in} }, Z) \\
\rightarrow (Y, \text{ \textit{located\_in} }, Z).
\end{multline}
Based on this rule, in Figure~\ref{fig:motiv}(b), we can induce the missing triplet $(Manchester United, \text{ \textit{located\_in} }, Manchester)$\newline 
due to the existence of the two-hop path consisting of the relations $(Cristiano, \text{ \textit{lives\_in} }, Manchester)$ and $(Cristiano, \text{ \textit{part\_of} } ,Manchester United)$.\footnote{Technically speaking, these methods only learn the ``and" operation between relations. We are interested in expanding to more sophisticated rules. But this is beyond the scope of this paper.}

Previous inductive relation prediction methods~\cite{galarraga2013amie, meilicke2018fine} mainly search for the good rules in a KG to predict the existence of a target triplet. However, the number of possible good rules is exponential to the size of the KG. In case of large searching cost, these works need to introduce artificial pruning methods, thus cannot suit many benchmarks. Inspired by the rapid development of deep learning, ~\citet{yang2017differentiable} and \citet{sadeghian2019drum} introduce neural networks to approximate the searching of rules. However, the approximation is rather coarse, leading to inferior performance in downstream tasks.
 
To avoid searching through the exponential-size space of rules, we tackle the problem from a new algebraic topology perspective. We view logical rules as cycles, and then learn good rules in the space of cycles. In fact, any rule can be considered a cycle by including both the relation path and the target relation itself, e.g., the red cycle in Figure \ref{fig:motiv}(a).

The benefit of using cycles is that there is an intrinsic algebraic structure in the space of cycles. Based on the mathematics of algebraic topology \citep{munkres2018elements}, \emph{the space of cycles is a vector space} under certain assumptions.\footnote{More specifically, we assume the coefficient is in $\mathbb{Z}_2$.} We can exploit the linear structure of the cycle space for efficient rule learning. In particular, we focus on a \emph{basis}, i.e., a set of linearly independent cycles that can be combined to represent any cycle.  
Taking Figure~\ref{fig:motiv}(a) as an example, if we choose the red cycle and the blue cycle as the cycle basis, the green cycle can be represented as their sum. Here ``sum'' means modulo-2 sum (Figure \ref{fig:motiv}(d)). 

In general, a cycle basis with $n$ cycles can represent $2^n-1$ possible cycles; any cycle can be uniquely written as a linear sum of the basis cycles with 0/1 coefficients. Here $n$ is called the \emph{Betti number of dimension 1}.  
In the example graph in Figure \ref{fig:CBGNN}(a), the Betti number is three. Each cycle basis (examples in Figure \ref{fig:CBGNN}(b)) has three cycles, but can represent all $2^3-1=7$ cycles.
By focusing on the cycle basis that spans the cycle space, we decrease the parameter space from exponential to linear.

We propose \textit{Cycle Basis Graph Neural Network (CBGNN)} to efficiently learn the good cycles (i.e., cycles representing good rules) through a given cycle basis. Our method searches through all possible combinations of the cycles in the basis, that is, the whole space of cycles.
We build a GNN on a new graph whose nodes represent cycles in the chosen basis, and edges represent their interaction. Through the message passing of the GNN, we are running implicit algebraic operations over the space of cycles. Our method will efficiently find good cycles from the whole space of cycles. These good cycles help predict triplets in a KG. 

One challenge is to choose a suitable cycle basis for the method. 
Considering that short rules can be learned more effectively~\cite{sadeghian2019drum,teru2020inductive}, we assume that suitable cycle bases should generally contain more short cycles for an effective encoding. Notice that we are not directly choosing the short cycles in the cycle basis, but learning the right cycles (even if it is long) with the cycle basis. Inspired by the theory of algebraic topology~\cite{dey2010approximating,chen2010measuring,busaryev2012annotating}, we exploit a set of shortest-path-tree cycle bases to guarantee sufficient coverage and shortness. Through thorough experiments on various benchmarks, we will show that the well selected cycle bases can learn the desired good cycles effectively and efficiently, thus solving the inductive relation prediction problem.

Our paper proposes a novel cycle-centric perspective of graph representation learning.
This is quite different from the popular directions such as node representation learning~\cite{kipf2016semi,ye2019curvature,zhao2020persistence, yan2021link} and graph representation learning~\cite{xu2019powerful, gao2019graph}. It is closer to the new trend of learning advance graph representations such as line graph representation~\cite{cai2021line} and simplicial complex representation~\cite{bodnar2021weisfeiler1}.  

\newpage

In summary, our contribution is three-fold:
\begin{enumerate}[topsep=1pt,itemsep=1pt,partopsep=1pt, parsep=1pt]
	\item We investigate, for the first time, the inductive relation prediction problem through a cycle-centric perspective. Unlike traditional methods, our model reasons and learns through the space of cycles to find good rules. 
	\item Inspired by the mathematics of algebraic topology, we propose to exploit the linear structure of the cycle space, and to compute suitable cycle bases that can best express the rules. This empowers us to explore rule space more efficiently than previous approaches.
	\item We propose a novel graph neural network, CBGNN. It runs implicit algebraic operations in the cycle space through the message passing of a GNN, and learns the representation of good rules. Experiments show that CBGNN achieves state-of-the-art results on various inductive relation prediction benchmarks.
\end{enumerate}

\section{Related Works}
\label{sec:related}
\textbf{Graph Learning with Topology.}
Graph structural information has been shown to enhance graph representation learning~\citep{kipf2016semi,you2019position,ye2019curvature}. In recent years advanced topological information, i.e., persistent homology \citep{edelsbrunner2000topological, edelsbrunner2010computational}, have been applied to graph learning problems. 
These features can provide additional discriminative power for various graph representation learning tasks \citep{hofer2020graph, carriere2020perslay, hofer2017deep, zhao2019learning, yan2021link,bhatia2019persistent, zhao2020persistence, yan2022neural}. 
From a different perspective, new graph neural networks have been proposed for high-order graphs, treated as simplicial or cell complexes \citep{bodnar2021weisfeiler1,bodnar2021weisfeiler2}. 
Beyond graph data, topological information has been used in many other learning contexts, such as in imaging \citep{hu2019topology,hu2021topology,wang2020topogan}, robust learning \citep{chen2019topological,wu2020topological,zheng2021topological}, biomedicine \citep{chan2013topology,rizvi2017single,aukerman2020persistent,wang2021topotxr}, neuroscience \citep{petri2014homological,giusti2015clique,li2017metrics,kanari2018topological}, etc. 

In topological data analysis, \emph{homology localization}, including computing short cycles representatives of a homology class and computing short cycle bases representing the whole homology group, is well studied theoretically \citep{chambers2009minimum,chen2011hardness,dey2011optimal,busaryev2012annotating,dey2010approximating,dey2022computational}. In recent years, new questions have been raised regarding finding short representative cycles for classes in persistent homology \citep{wu2017optimal,dey2020computing}. 
Inspired by these works, we exploit the space of cycles and its underlying algebraic structure for better graph representation learning. We believe the cycle-centric design of our graph neural network is generic and can extend to many other tasks beyond relation prediction.

\textbf{Inductive relation prediction methods.} Inductive relation prediction methods can be divided into two categories: path-based methods, and GNN-based methods. 
Path-based methods mainly view rules as paths, i.e., sequences of relations connecting two entities of interest. Among path-based methods, AMIE~\citep{galarraga2013amie} and RuleN~\citep{meilicke2018fine} are classic rule learning methods. These two methods prune the process of rule searching based on strong assumptions on the attribute of rules, thus their performances are not satisfying. NeuralLP \citep{yang2017differentiable} and DRUM \citep{sadeghian2019drum} learn a weight for each relation type, and then weigh a path with the product of the weights of its relations. This approximation, although reduces the number of parameters from exponential to linear, is rather coarse and results in unsatisfying performance.

GNN-based methods such as GraIL~\citep{teru2020inductive} and CoMPILE~\citep{mai2020communicative} predict missing triplets with graph neural networks (GNNs). 
To predict whether a certain triplet exists in the KG, these methods first extract the corresponding vicinity graph of the triplet and then learn the rules through message passing and GNN scoring. Therefore, they can only predict the triplets one by one, with a rather low computational efficiency. 

In addition, existing inductive relation prediction methods \citep{yang2017differentiable, sadeghian2019drum, teru2020inductive, mai2020communicative} limit the length of learned rules to a small number in case of high computational cost, 
while our framework can represent long rules with short cycles. We will empirically show the efficiency and the effectiveness of our framework by comparing it with these models.

\section{Cycle Space, Cycle Basis, and the Pursuit of Suitable Bases}
\label{sec:pre}
In this section, we explain how to find suitable cycle bases that can facilitate the learning of good cycles/rules. We first introduce the background of the cycle space and cycle basis. Next, we explain our choice of suitable cycle bases, which will be the foundation of our model.

By no means our exposition is comprehensive. For more details, we refer the readers to textbooks on algebraic topology and computational topology \citep{munkres2018elements,edelsbrunner2010computational,dey2022computational}. 
We focus on cycles in undirected graphs, while the definitions generalize to higher dimensions, e.g., simplicial complexes. Furthermore, we focus on the algebraic structures over $\mathbb{Z}_2$ field, which has two elements, $0$ and $1$, under modulo-2 addition and multiplication. Over $\mathbb{Z}_2$ field, the structure of the space of cycles is simpler and more friendly to computation.

For the rest of the paper, regarding the input KG, we will use node, vertex, and entity interchangeably. We will also use edge and triplet interchangeably.
Within this section, we temporarily ignore the relation associated with each triplet. We treat the input KG as an undirected graph $G=(V,E)$, where $V$ and $E$ denote the sets of vertices and edges.

\subsection{Background: Cycle Space and Cycle Basis}
For ease of exposition, we assume the input graph $G$ is connected. The definitions can easily extend to a graph with multiple connected components.
An \emph{elementary cycle} is a closed loop, i.e., a sequence of edges, $\{(v_0,v_1),(v_1,v_2),\ldots,(v_{n-1},v_n), (v_n,v_0)\}$, going through distinct vertices except for the first and the last. A \emph{cycle} $z$ is the union of a set of elementary cycles. 

The set of all cycles constitute a vector space under modulo-2 additions and multiplications. Figure \ref{fig:motiv}(d) illustrates the mod-2 addition of cycles.
There is a nice linear algebra interpretation of the space of cycles. Assume a fixed indexing of all edges and all vertices. The $|V|\times|E|$ incidence matrix, $\partial$, also called the \emph{boundary matrix},  encodes the adjacency relationship between edges and vertices. Any set of edges, called a \emph{chain}, corresponds to an $|E|$-dimensional binary vector, $c$. The $i$-th entry of $c$, $c_i$, is 1 if and only if the chain contains the $i$-th edge, $e_i$. The set of all chains form a vector space called the \emph{chain group}. All chains one-to-one correspond to all possible $|E|$-dimensional binary vectors. Multiplying the boundary matrix to a given chain is equivalent to taking the boundary of the chain. Figure \ref{fig:motiv}(c) and (e) show a sample graph and its boundary matrix. A \emph{cycle} is a chain with zero boundary. Formally, the set of all cycles of $G$, denoted as $\mathcal{Z}_G$, is the kernel space of the boundary matrix, $\mathcal{Z}_G = \ker \partial = \{c\mid \partial c = 0\}$. In the example graph in Figure \ref{fig:motiv}(c), there are 3 different nontrivial cycles, highlighted in red, blue, and green\footnote{Technically, an empty chain (contains no edges) is also a cycle.}.

\myparagraph{Cycle basis.}
A \emph{cycle basis} is a basis spanning the cycle space $\mathcal{Z}_G$.
Formally, a basis, $Z$, is a maximal set of cycles $\{z_1,z_2,\ldots\}$ such that (1) any cycle in $\mathcal{Z}_G$ can be written as the formal sum of cycles in the basis, $\forall z\in \mathcal{Z}_G, \exists \alpha_i\in \{0,1\}, s.t. \text{ } z=\sum\nolimits_{z_i\in Z} \alpha_i z_i$  and (2) cycles in $Z$ are linearly independent, 
$\sum\nolimits_{z_i\in Z} \alpha_i z_i = 0 \iff \forall z_i \in Z, \alpha_i = 0$.
In Figure \ref{fig:motiv}(c), the red and the blue cycles form a cycle basis. We note that the basis is not unique. The red cycle and the green cycle form another cycle basis of the same graph. However, the number of elements in the basis, $|Z|$, is the same. We call it the \emph{Betti number}, denoted as $\beta$. We have $\beta=|E| - |V| + 1$, and the cycle space has size $2^\beta$.

\subsection{The Pursuit of Suitable Cycle Bases}
\label{subsec:pursuit}

The central idea of our approach is to find practical and efficient cycle bases to represent the cycle space, so that we can efficiently learn any "good cycle" in the graph which possibly corresponds to a good rule. 
In this section, we explain how such cycle bases are constructed.
In theory, any basis can represent the whole cycle space, and thus can serve the purpose. However, during learning, we look for a practically suitable basis or a set of suitable bases that can easily represent any good cycle. We expect the cycle bases to meet the following criteria: (1) cycles in the bases can be easily encoded for feature representation; (2) any cycle in the KG can be easily represented by cycles in the bases.

\textbf{Cycle bases that can be easily encoded.}
To have a better learning performance, we need to encode cycles into feature representations. 
To ensure an effective encoding, we prefer bases with short cycles. Note that this does not exclude long good cycles from being found. Our method essentially finds combinations of cycles from the chosen bases to form (potentially long) good cycles. And the feature representations of the good cycles are derived from the features of its relevant cycles in the bases.

Motivated by this, we represent good cycles using \emph{shortest path tree (SPT) cycle bases}, i.e., cycle bases constructed based on shortest path trees. They generally contain relatively short cycles, and can be computed efficiently~\citep{dey2022computational,dey2010approximating,chen2010measuring}. 

\begin{figure*}[hbtp]
	\centering
	\includegraphics[width=0.9\textwidth]{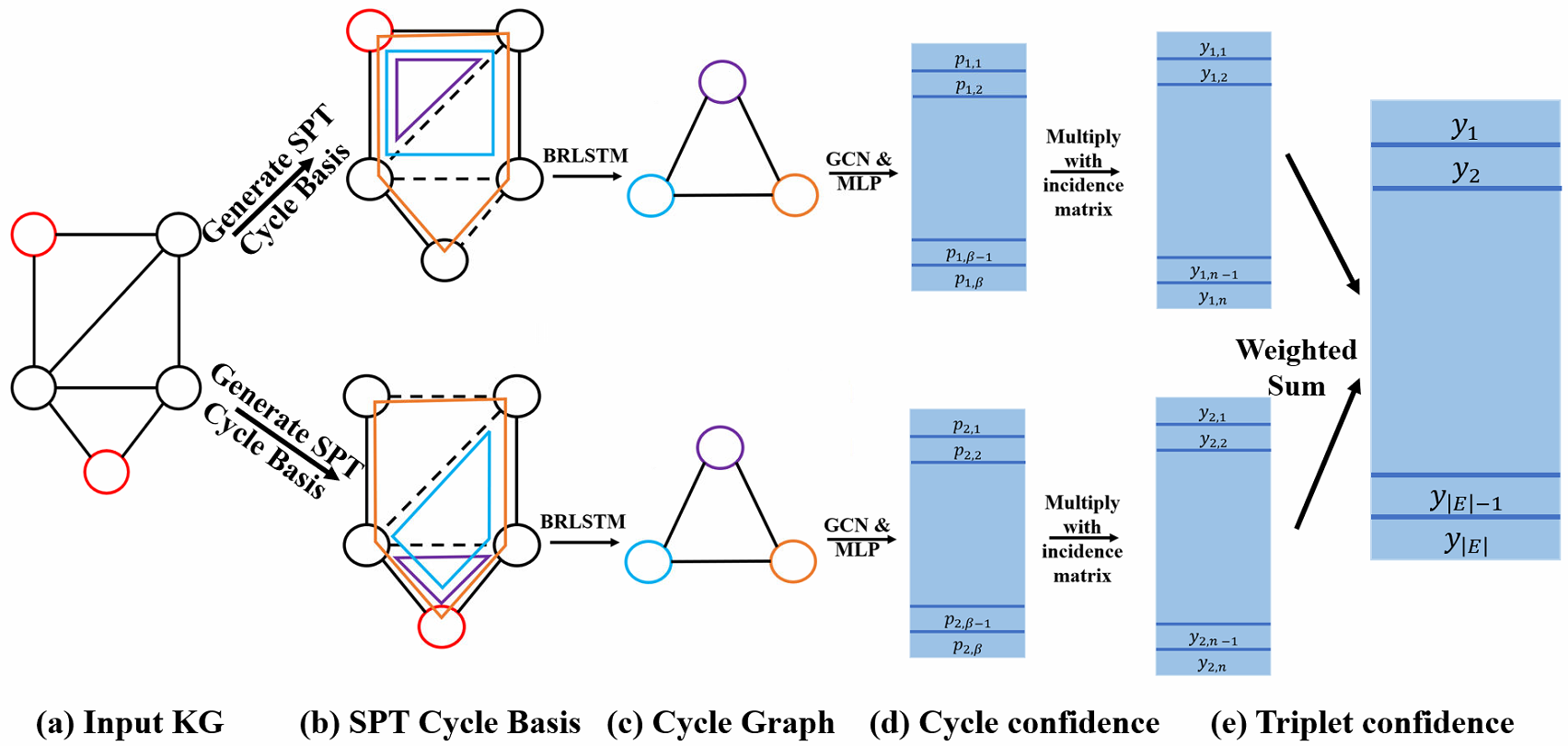}
	\vspace{-.1in}
	\caption{Architecture of CBGNN. (a) the input KG with the selected red root nodes. (b) the SPT cycle bases from (a). (c) the cycle graph where nodes represent cycles in (b) and edges indicate a strong interaction between cycles. The initial node feature vectors are extracted by BR-LSTM. (d) the confidence value of cycles in (c). (e) the confidence value of target triplets, and we use the weighted sum of triplet confidence from different SPT cycle bases as the final triplet confidence.} 
	\label{fig:CBGNN}
	\vspace{-.1in}
\end{figure*}

Formally, a \emph{shortest path tree (SPT)} is a spanning tree $T_p\subseteq G$ with root $p$, such that for any vertex $q\neq p$, its path to $p$ within $T_p$ is also its shortest distance path to $p$ within $G$. In other words, $T_p$ is a union of shortest paths from all vertices to the root $p$. A shortest path tree defines a unique cycle basis, which we call the \emph{SPT cycle basis}. 
As shown in Figure \ref{fig:CBGNN}(a) and (b), given a shortest path tree, $T_p$, each non-tree edge $e\in E\backslash T_p$ forms an elementary cycle with the tree $T_p$. 
We construct the basis by enumerating through all non-tree edges and collect all the corresponding elementary cycles. We denote this cycle basis $Z(T_p)$.
An SPT cycle basis naturally contains short cycles; each cycle is a composition of an edge $(u,v)$ and two shortest paths - the shortest path from $u$ to $p'$ and the shortest path from $v$ to $p'$. Here $p'$ is the lowest common ancestors of $u$ and $v$ within the rooted tree, $T_p$. 

\textbf{Cycle bases that can efficiently represent good cycles.} Given a single SPT cycle basis, a cycle that is away from the root are hard to be represented; it potentially requires many cycles from the given basis to represent. To efficiently represent all possible good cycles,
we collect a family of SPT cycle bases with different tree roots to ensure locality and sufficient coverage. These bases complement each other and achieve the best learning efficiency in representing good cycles. The hope is that any good cycle can be easily represented by at least one of the bases. 

Ideally, we can use the whole vertex set $V$ as roots and build the collection of cycle basis $\{Z(T_p)\mid p\in V\}$. This family of bases has been shown to have theoretical benefit \citep{dey2010approximating,chen2010measuring}. 
In practice, we cannot afford to construct the bases using all vertices as the SPT roots. We propose to sample vertices that are generally far away from each other. We perform spectral clustering on the graph and use centers of the clusters as the sample vertices, $S$. We hypothesize that these SPT cycle bases will cover the whole graph, and their corresponding cycle bases, $\{Z(T_p)\mid p\in S\}$, will satisfy our needs. We call these bases the \emph{SPT cycle bases family}. As validated in the appendix, these SPT cycle bases provide sufficient locality and coverage of the target edges/triplets, with short cycle representations, compared with random cycle bases.

\section{Cycle Basis Graph Neural Network (CBGNN)}
\label{sec:CBGNN}
In this section, we describe how to use the SPT cycle bases family to learn a good cycle representation, to find good rules, and to predict the existence/non-existence of a triplet. We propose a novel GNN based on the cycle bases, called CBGNN. 
The input of CBGNN is a KG and the target triplets. A target triplet $(e_h,r,e_t)$ refers to a query of whether the relation $r$ exists between entities $e_h$ and $e_t$. A target triplet is labeled positive if it exists in the KG, and negative otherwise.
Following the tradition \citep{teru2020inductive, mai2020communicative}, we temporarily add the negative triplets into the input graph.
CBGNN learns representations of cycles that best serve the goal of relation prediction and output the confidence value of the target triplets. 

The overview of our method is shown in Figure~\ref{fig:CBGNN}. Our method has two phases. In the first phase, we construct the cycle bases and build a new graph for each cycle basis (called the \emph{cycle graph}). 
In the cycle graph, nodes represent cycles in the basis, and nodes are connected if their corresponding cycles have a strong interaction. The information of the cycles can be converted into node features in the new graph for the next phase. Details are provided in Section~\ref{subsec:tree}.

In phase two (Section~\ref{subsec:GNN}), we build a GNN on the cycle graph to learn the confidence value for cycles. The confidence values for cycles are mapped to the confidence values for target triplets.
We construct GNNs for different cycle bases. These GNNs share weights and their aggregation is used to predict the confidence value for the target triplets.

\subsection{Generating the Cycle Graphs}
\label{subsec:tree}

Recall that in Section~\ref{sec:pre}, we sample vertices at different parts of the input KG and construct SPT cycle bases accordingly.
In order to achieve good locality and coverage, these vertices should be selected sufficiently apart from each other. In this way, the family of cycle bases can effectively represent all cycles at different parts of the input KG. 
In particular, we run spectral clustering on the input graph and partition the nodes into $k$ clusters. Then we take 
the node closest to the cluster center as the set of sample vertices, $S$. 

Using vertices in $S$ as roots, we use the breadth-first-search algorithm to construct $k$ SPTs. The complexity for building each SPT is $O(|V|+|E|)$. \footnote{Note the breadth-first-search algorithm works only because we assume all edges are weighed one. }
For each SPT, $T_p$, $p\in S$, we construct its cycle basis by going through all non-tree edges. For each non-tree edge, $(u,v)\in E\backslash T_p$, we find the least common ancestor of $u$ and $v$ in $T_p$ in $O(|V|)$ time. In total the complexity for building one cycle basis is $O(|E||V|)$ All the cycles form the desired SPT cycle basis $Z(T_p)$. We now have $k$ cycle bases, each of which has $\beta$ many cycles. The total running time for building $k$ bases is $O(k|V||E|)$. 

For input graphs which consist of several connected components, the cycle bases of different component graphs are independent of each other. We treat the component graphs as separate input graphs, and generate $k$ SPT cycle bases for each of them. We essentially construct a CBGNN for each component graph, although their weights are all shared. 

\myparagraph{Cycle incidence matrix.} We explicitly construct a \emph{cycle incidence matrix} for each SPT cycle basis. This matrix encodes the incidence relationship between cycles and edges in the input KG. 
It will be used at different stages of our learning
because it can provide a convenient way to map confidence values between cycles and triplets.

For each of the $k$ constructed cycle bases, we construct the cycle incidence matrix $C_T$ as an $|E|\times \beta$ binary matrix. Each column corresponds to one cycle in the basis. Each row corresponds to an edge/triplet in the input graph. The $(i,j)$-th entry of the matrix is 1 if the $j$-th cycle contains the $i$-th edge, and 0 otherwise. An edge may not be associated with any of the basis cycles and thus has all zeros in the corresponding row. 
See Figure \ref{fig:motiv}(f) for an illustration. After generating $k$ shortest path trees and their SPT cycle bases, we acquire $k$ cycle incidence matrices: $\{C_T^1, C_T^2, ..., C_T^k\}$.

\myparagraph{Cycle feature.} 
To use these cycles in learning, we need to extract their attributes/features.
We need a feature representation for a cycle based on the relations associated with its triplets. 
Inspired by existing methods on rule learning~\citep{marcheggiani2017encoding, vashishth2019composition}, we propose a recurrent model, \textit{Bi-Relational LSTM (BR-LSTM)}, as the feature generator. It converts a cycle of triplets into a fixed-length feature vector for the CBGNN to use.

We assume that information in an edge/triplet flows along both directions, and encode the cycle in a relation-aware manner.
We denote by $(u, r, v)\in E$ a triplet connecting nodes $u$ and $v$ with relation $r$. Here $E$ is the set of all the triplets, we assume that an inverse triplet $(v, r^{-1}, u)$ is also included in the KG. $r^{-1}$ is defined as the inverse relation of $r$. 
Formally, we extend the triplet set of the KG as: $E' = E \cup \{(v,r^{-1},u)| (u,r,v) \in E\}$. An illustration of the construction can be found in the appendix.

For each cycle, we can use LSTM to encode the cycle from both directions using triplets in $E'$. Take Figure~\ref{fig:motiv}(a) as an example, for simplicity we substitute the relations \textit{part\_of}, \textit{lives\_in}, and \textit{located\_in} with $r_1$, $r_2$ and $r_3$, respectively. The rule can be represented by $(Lebron, r_1, Lakers) \wedge (Lebron, r_2, L.A) \rightarrow (Lakers, r_3, L.A)$. In practice, we use the non-tree edge (target triplet) in the cycle as the first triplet of the sequence. Therefore we convert the rule into two sequences with the opposite direction: $(Lakers, r_3, L.A), (L.A, r_2^{-1}, Lebron) , (Lebron, r_1, Lakers) $ and $(L.A, r_3^{-1}, Lakers), (Lakers, r_1^{-1},LeBron)$, $(Lebron, r_2, L.A)$. We denote the two sequence as $s_1$ and $s_2$. To encode the two sequences, we adopt a LSTM for each sequence, to capture the contextual information between relations:
$$w_{next_1}, (h_{s_1}, c_{s_1}) = LSTM(w_{s_1}, (h_1, c_1));$$
$$w_{next_2}, (h_{s_2}, c_{s_2}) = LSTM(w_{s_2}, (h_2, c_2)).$$
Here, for any $i = 1, 2$, $w_{s_i}$ denotes the input embedding vector for sequence $s_i$. $h_i$ and $c_i$ are the initial hidden state and cell state for sequence $s_i$, they are initialized as zero. $w_{next_i}$ is the output features from the last layer of the LSTM. It is not needed in our setting. $h_{s_i}$ and $c_{s_i}$ are output hidden state and cell state for the whole sequence $s_i$. We use them as the feature vector for each sequence. The final feature vector for the rule and its corresponding cycle, $z$, is $x_z = (h_{s_1} + h_{s_2}) \bigoplus (c_{s_1} + c_{s_2})$, where $\bigoplus$ represents the concatenation of vectors. 

\begin{table*}[hbtp]
		\centering
		
	\vspace{-0.2 in}	\caption{{AUC-PR scores of inductive relation prediction, the baseline results are copied from \citep{teru2020inductive, mai2020communicative}.} }
		\label{tab:expres} 
		\scalebox{0.70}{
			\begin{tabular}{|l|cccc|cccc|cccc|}
				\hline\noalign{\smallskip}
				 &\multicolumn{4}{c|}{WN18RR} &\multicolumn{4}{c|}{FB15K-237}&\multicolumn{4}{c|}{NELL-995} \\
				 \noalign{\smallskip}\hline\noalign{\smallskip}
				Method & v1 & v2 & v3 & v4 & v1 & v2  & v3 & v4&  v1 & v2 & v3 & v4\\
				\noalign{\smallskip}\hline\noalign{\smallskip}
				NeuralLP & 86.02 & 83.78 & 62.90 & 82.06 & 69.64 & 76.55 & 73.95 & 75.74 & 64.66 & 83.61 & 87.58 & 85.69\\
				DRUM & 86.02 & 84.05 & 63.20 & 82.06 & 69.71 & 76.44 & 74.03 & 76.20 & 59.86 & 83.99 & 87.71 & 85.94\\
				RuleN & 90.26 & 89.01 & 76.46 & 85.75 & 75.24 & 88.70 & 91.24 & 91.79 & 84.99 & 88.40 & 87.20 & 80.52\\
				GraIL & 94.32  & 94.18 & 85.80 & 92.72 & 84.69 & 90.57 & 91.68 & 94.46 & \textbf{86.05} & 92.62 & 93.34 & 87.50\\
				CoMPILE & 98.23 & \textbf{99.56} & \textbf{93.60} & \textbf{99.80} & 85.50 & 91.68 & 93.12 & 94.90 &  80.16 &  \textbf{95.88} & 96.08 & 85.48\\
				\noalign{\smallskip}\hline\noalign{\smallskip}
				CBGNN & \textbf{98.63} & 97.62 & 89.76 & 97.80 & \textbf{96.34} & \textbf{96.53} & \textbf{96.38} & \textbf{95.23} & 82.79
				& 94.78 & \textbf{96.29}& \textbf{94.02}\\
				\noalign{\smallskip}\hline
		\end{tabular}}
		\vspace{-0.2 in}
	\end{table*}

	\begin{table*}[hbtp]
		\centering
		\caption{Hit@10 scores of inductive relation prediction, the baseline results are copied from \citep{teru2020inductive, mai2020communicative}.}
		\label{tab:hits} 
		\scalebox{0.73}{
			\begin{tabular}{|l|cccc|cccc|cccc|}
				\hline\noalign{\smallskip}
				&\multicolumn{4}{c|}{WN18RR} &\multicolumn{4}{c|}{FB15K-237}&\multicolumn{4}{c|}{NELL-995} \\
				\noalign{\smallskip}\hline\noalign{\smallskip}
				Method & v1 & v2 & v3 & v4 & v1 & v2  & v3 & v4&  v1 & v2 & v3 & v4\\
				\noalign{\smallskip}\hline\noalign{\smallskip}
				NeuralLP & 74.37 & 68.93 & 46.18 & 67.13 & 52.92 & 58.94 & 52.90 & 55.88 & 40.78 & 78.73 & 82.71 & 80.58\\
				DRUM & 74.37 & 68.93 & 46.18 & 67.13 & 52.92 & 58.73 & 52.90 & 55.88 & 19.42 & 78.55 & 82.71 & 80.58\\
				RuleN & 80.85 & 78.23 & 53.39 & 71.59 & 49.76 & 77.82 & 87.69 & 85.60 & 53.50 & 81.75 & 77.26 & 61.35\\
				GraIL & 82.45 & 78.68 & 58.43 & 73.41 & 64.15 & 81.80 & 82.83 & 89.29 & 59.50 & 93.25 & 91.41 & 73.19\\
				CoMPILE & 83.60 & 79.82 & 60.69 & 75.49 & 67.64 & 82.98 & 84.67 & 87.44 & 58.38 & 93.87 & 92.77 & 75.19\\
				
				\noalign{\smallskip}\hline\noalign{\smallskip}
				CBGNN & \textbf{98.40} & \textbf{96.14}  & \textbf{62.28} &\textbf{96.50} & \textbf{97.56}  & \textbf{96.03} & \textbf{94.91} & \textbf{94.73} &  \textbf{84.00}
				& \textbf{94.96} & \textbf{95.34} & \textbf{92.34}\\
				\noalign{\smallskip}\hline
		\end{tabular}}
		\vspace{-0.2 in}
	\end{table*}
	
\subsection{GNN Learning with Cycle Graphs}
\label{subsec:GNN}
We propose a GNN to exploit the SPT cycle bases to learn representations of good rules and use the learned rules to predict the confidence value of certain triplets. We first build the cycle graphs for the SPT cycle bases, and then learn the confidence value for cycles and triplets.

\textbf{Building cycle graphs.} Recall that in Section~\ref{subsec:tree}, we obtain $k$ cycle bases and $k$ corresponding $C_T$ matrices. For each cycle basis, we construct a new graph in which nodes represent cycles in the cycle basis and edges indicate that the two corresponding cycles have a strong interaction. To measure the interaction between any two cycles in the basis, we compute their overlapping, i.e., the number of triplets they share. In the new graph, each cycle is connected with its top $m$ overlapping neighbors, i.e., the top $m$ other cycles with the most number of shared triplets.
To compute the number of shared triplets between all pairs of cycles in the basis, we simply multiply the cycle incidence matrix and its transpose, $C_T^T\cdot C_T$, and read the entries of the resulting $\beta\times\beta$ matrix.

\textbf{Learning cycle representation and confidence.} To learn the representation and confidence values of the desired rules, we apply a classic $L$-layer graph convolutional network (GCN) ~\citep{kipf2016semi} to the constructed cycle graph. The input is the feature vector of cycles generated by BR-LSTM, and the output is the representation of cycles. The message passing by GCN drives the information flow between different cycles. After an $L$-layer GCN, the embedding of a certain node can be viewed as a combination of node representation from its $L$-hop neighborhood. Take Figure~\ref{fig:motiv} as an example, if we take $\{z_r, z_g\}$ as the cycle basis, then $z_r$ and $z_g$ will be the nodes in the new graph. Because they share triplets $e_1$ and $e_4$, there is an edge between the two nodes in the new graph. Then cycle $z_b$ can be learned by the message passing between $z_r$ and $z_g$. 

In the $\ell$-th layer of GCN, we can obtain the embedding matrix $X^{\ell} = [x_1^{\ell}, x_2^{\ell},...,x_{\beta}^{\ell}]$ where $x^{\ell}_i \in \mathbf{R}^{d_{\ell}}$ is the representation of node $i$ in the $\ell$-th layer, $\ell=0,1,...,L$. Here, $X^0$ is the initial cycle features from BR-LSTM, and $X^{L}$ is the node embedding matrix of the final layer. After the $L$-layer GCN, we adopt a two-layer Multi-Layer Perceptron (MLP) followed by a sigmoid function to learn the confidence value for each cycle in the basis: $P = sigmoid(MLP(H^L))$, where $P = [p_1, p_2,...,p_{\beta}] \in R^{\beta}$, $0 \leq p_i \leq 1$. See Figure \ref{fig:CBGNN}(c) and (d) as an illustration.

\textbf{Learning triplet confidence.} Finally, we compute the confidence values for the triplets of KG based on the confidence values of cycles learned through GNN.
We take the max confidence value of cycles/rules that pass a triplet as the confidence value for the triplet.
Recall that for each cycle basis, the cycle incidence matrix $C_T$ stores the incidence relationship between cycles and triplets. The $i$-th row of matrix $C_T$ has 1's corresponding to cycles in the basis that pass triplet $e_i$. 
For triplet $e_i$, its confidence value is computed as $y_i = max(C_T(i,\ast) \odot P)$, where $C_T(i,\ast)$ is the $i$-th row of $C_T$, and $\odot$ denotes the element-wise product between two vectors. We obtain the confidence values for all target triplets: $Y = [y_1, y_2, ..., y_{n}] \in R^{n}$, where $n$ is the total number of target triplets, $0 \leq y_i \leq 1, i = 1,2,...,n$.
We aggregate the output of the $k$ GCNs to obtain the final triplet confidence. Each GCN is built on one SPT cycle basis and its corresponding cycle graph. We compute the final confidence value of each triplet using a weighted sum of the triplet confidence from different GCNs. 
Formally, $Y_{final} = \sum_{i = 1}^{k} w_iY_i / \sum_{i = 1}^{k}w_i$.
We train CBGNN by minimizing the cross-entropy loss on target triplets.

\section{Experiments}
\label{sec:exp}
	
\begin{table*}[hbtp]
		\centering
	\vspace{-0.2 in}	\caption{{Evaluation of computational efficiency (second).} }
		\label{tab:time} 
		\scalebox{0.68}{
			\begin{tabular}{|l|ccc|ccc|ccc|}
			
				\noalign{\smallskip}\hline\noalign{\smallskip}
				Dataset & \multicolumn{3}{c|}{WN18RR v1}  & \multicolumn{3}{c|}{FB15K-237 v1}  &  \multicolumn{3}{c|}{NELL-995 v1}   \\
			    Phase & Preparation & Training & Inference &  Preparation & Training & Inference &  Preparation & Training & Inference  \\
			    \noalign{\smallskip}\hline\noalign{\smallskip}
			    GraIL & 452.36 & 2230.55 & 1.07 & 704.42 & 9026.21 & 1.67   &  402.86 & 3718.22 &  1.79\\  
			    CoMPILE & 434.45 & 2388.28 & 1.46 & 706.19 & 3809.56 & 2.41 & 479.21 & 2868.38 & 1.23\\
			    CBGNN & 601.96 &  952.55 & 0.52 & 437.13 & 901.27 & 0.75 & 379.29 & 175.19 & 0.14\\
				\noalign{\smallskip}\hline
		\end{tabular}}
		\vspace{-0.1 in}
	\end{table*}
\begin{table*}[hbtp]
		\centering
	\vspace{-0.1 in}	\caption{{AUC-PR scores of ablation study.}}
		\label{tab:ablation} 
		\scalebox{0.65}{
			\begin{tabular}{|l|cccc|cccc|cccc|}
				\hline\noalign{\smallskip}
				&\multicolumn{4}{c|}{WN18RR} &\multicolumn{4}{c|}{FB15K-237}&\multicolumn{4}{c|}{NELL-995} \\
				\noalign{\smallskip}\hline\noalign{\smallskip}
				Method & v1 & v2 & v3 & v4 & v1 & v2  & v3 & v4&  v1 & v2 & v3 & v4\\
				\noalign{\smallskip}\hline\noalign{\smallskip}
		   \textcolor{black}{CBGNN-MLP} & 96.33  & 97.49  & 86.86 & 95.22 & 90.90 & 94.07 & 87.01 & 87.78  & 72.29 & 93.35 & 94.63 & 91.29\\
		   
		   \noalign{\smallskip}\hline\noalign{\smallskip}
		   CBGNN-Random & 97.13  &  76.64 & 87.30 & 93.47 & 96.23 & 96.07 & 93.27 & 94.49  & \textbf{83.69} & 93.73 & 96.20 & 92.94\\
				CBGNN-Single &  58.96 & 58.05 & 55.67  & 61.61 & 81.67 & 84.28  &81.75  & 79.44 & 72.29 & 83.79 & 90.70 & 80.97\\		
				\noalign{\smallskip}\hline
				CBGNN-BOW & 97.54 & 96.45 & 86.83 & 97.46 & 96.04 & \textbf{97.61} & \textbf{96.85} & \textbf{97.00} & 75.31 & 90.25 & 91.00 & 87.53\\
				CBGNN-LSTM & 98.26 & 97.04 & 89.69 & 97.75 & 95.86 & 91.46 & 94.56 & 92.47 & 71.85 & 93.33& 93.74 & 85.78\\
				\noalign{\smallskip}\hline
				CBGNN & \textbf{98.63} & \textbf{97.62} & \textbf{89.76} & \textbf{97.80} & \textbf{96.34} & 96.53 & 96.38 & 95.23 & 82.79
				& \textbf{94.78} & \textbf{96.29}& \textbf{94.02}\\
				\noalign{\smallskip}\hline
		\end{tabular}}
		\vspace{-0.2 in}
	\end{table*}
	
We compare our methods with state-of-the-art (SOTA) inductive relation prediction models on popular benchmark datasets. We also use ablation studies to demonstrate the efficacy of different proposed modules in our method. Further experimental details can be found in the appendix. The code is provided in \href{https://github.com/pkuyzy/CBGNN}{https://github.com/pkuyzy/CBGNN}.

\textbf{Datasets.}
We use SOTA benchmark datasets proposed in \citep{teru2020inductive,mai2020communicative}. For inductive relation prediction, the entities in the training set and the test set should not be overlapped. Therefore the training and test sets are totally disjoint graphs. Details are provided in the appendix. Among these datasets, FB15k-237 has $>$ 200 relation types, NELL-995 contains an average of 50 relation types, and WN18RR contains $\approx$ 10 relation types. 

\textbf{Baseline.}
We compare with  SOTA inductive relation prediction methods including (1) path-based methods: NeuralLP~\citep{yang2017differentiable}, RuleN~\citep{meilicke2018fine}, DRUM~\citep{sadeghian2019drum} and (2) GNN-based methods: GraIL~\citep{teru2020inductive}, CoMPILE~\citep{mai2020communicative}. 

\textbf{Evaluation.} Similar to \citep{teru2020inductive, mai2020communicative}, we use area under the precision-recall curve (AUC-PR) and Hits@10 scores as the evaluation metrics. To calculate AUC-PR, we sample an equal number of non-existent triplets as the negative samples. To evaluate the Hits@10 score, we rank each positive triplet among 50
randomly sampled negative triplets. We run each experiment five times with different negative samples and report the mean results. 

\textbf{Negative sampling.} Following~\citep{teru2020inductive, mai2020communicative}, we sample negative triplets by replacing the head (or tail) of a true triplet with a randomly sampled entity. 


\textbf{Results and discussion.}
Table~\ref{tab:expres} and Table~\ref{tab:hits} show the AUC-PR scores and Hits@10 scores respectively. Our method outperforms all SOTA baselines in terms of Hits@10 (Table~\ref{tab:hits}). 
As for AUC-PR (Table~\ref{tab:expres}), our method outperforms nearly all SOTA baselines on FB15K-237 and NELL-995. On WN18RR, CBGNN is a close second, trailing marginally behind CoMPILE, but outperforming the remaining methods significantly.
Note that in terms of the number of relationship types, FB15k-237 ($>$200) and NELL-995 ($\approx$50) are significantly larger than WN18RR ($\approx$10). They are considered much more semantically complex. \emph{This demonstrates that our novel cycle-based approach has stronger modeling power for KGs with complex semantics.}

\textbf{Computational efficiency.} 
For all methods, we set the training epochs to 100 and run 5 times to report the average time. In Table~\ref{tab:time}, ``Preparation" denotes the time to extract subgraphs for GraIL and CoMPILE, and the time to generate 20 SPT cycle bases for CBGNN. ``Training" and ``Inference" denote the time of training 100 epochs and inference once respectively. As shown in the table, our method is significantly faster than existing GNN-based methods. 
For each training/testing triplet, existing GNN-based methods extract a subgraph within the vicinity and then apply graph convolution. Repeating over all target triplets is rather expensive in practice. On the contrary, our method construct one unified GNN for all target triplets and learn/predict their confidence values simultaneously.

\textbf{Ablation studies.}
We perform ablation studies to validate the efficacy of different proposed modules in CBGNN. We focus on three perspectives, the necessity of learning the needed cycles, 
the choice of cycle basis generation and the cycle feature generation. To show the necessity of learning the needed cycles with the SPT cycle bases, we compare with a baseline without GNN based on the cycle graph, called \emph{CBGNN-MLP}. CBGNN -MLP does not use a GNN model to perform the operation between cycles in the cycle bases, but directly uses the BRLSTM followed by a two-layer MLP to get the confidence of cycles. In this way, we are directly choosing the cycles from the SPT cycle bases rather than learning the needed rules as CBGNN does.
To justify the usage of SPT cycle bases, we compare with a baseline using randomly generated SPTs to build cycle bases, called \emph{CBGNN-Random}. Both CBGNN and CBGNN-Random generate the same number of trees/cycle bases, $k=20$. To show that sampling multiple trees/cycle bases is necessary, we also add a baseline with a single SPT cycle basis, called \emph{CBGNN-Single}.

For the generation of feature vectors for cycles in the cycle bases, we compare with two baselines which replace BR-LSTM with a bag-of-words-like (BOW) feature vector and a classic LSTM. These method are named \emph{CBGNN-BOW} and \emph{CBGNN-LSTM}, respectively. The BOW feature generates a histogram of different relation types within a given cycle. The classic LSTM takes a single direction to traverse through the loop instead of two. 

Results of the baselines are compared with the proposed CBGNN in Table~\ref{tab:ablation}.
In terms of the necessity of learning the needed cycles, CBGNN-MLP consistently performs worse than our method. This shows that the needed rules are not always in the SPT cycle bases, therefore we need to learn the right rules with the cycle bases rather than directly choosing from the SPT cycle bases. In addition, the result of CBGNN-MLP is comparable with state-of-the-art inductive relation prediction methods, showing that the SPT cycle bases contain a number of right rules. This provides empirical observation that SPT cycle bases are generally ``suitable cycle bases". 
In terms of cycle bases generation, our method generally outperforms CBGNN-Random. This demonstrates that in most cases, the center nodes of clusters are spread out and are capable of covering the whole graph. Thus for node selection, a clustering algorithm performs much better than random selection.
In addition, our method also outperforms CBGNN-Single, showing the necessity to utilize multiple bases to provide better coverage. In the appendix, we will provide more experiments on the influence of the number of SPT cycle bases $k$ on learning cycle representations.

In terms of cycle feature generation, our method outperforms CBGNN-BOW and CBGNN-LSTM on the majority of datasets. The results elucidate the efficacy of our relation-aware feature generation method, BR-LSTM. We were a bit surprised to find that BOW performs well on FB15k-237 and is slightly better than the proposed BR-LSTM. This may be due to the high semantic complexity of this dataset ($>$200 relationship types). The high number of relationship types makes LSTM and BR-LSTM hard to train, whereas BOW may perform robustly under such circumstances.

\section{Conclusion}
\vspace{-0.1 in}
We provide a novel GNN-based method for inductive relation prediction in knowledge graphs, and propose a cycle-centric approach that treats rule learning as a cycle learning problem for the first time. We exploit the intrinsic linear structure of the space of cycles and learn suitable cycle bases to represent the rules. The learning of cycle representation is carried out via a GNN that passes messages between cycles instead of nodes. Our approach achieves SOTA performance on various inductive relation prediction benchmarks, and provides a novel perspective in incorporating advanced topological information into graph representation learning. Also, our method can naturally be extended to tasks beyond relation prediction.

\textbf{Acknowledgements.}
We thank anonymous reviewers for their constructive feedback. This work is supported by the project of National Natural Science Foundation of China (No. 61876003), which is also a research achievement of Key Laboratory of Science, Technology and Standard in Press Industry (Key Laboratory of Intelligent Press Media Technology).

\nocite{langley00}

\bibliography{example_paper}

\begin{thebibliography}{65}
\providecommand{\natexlab}[1]{#1}
\providecommand{\url}[1]{\texttt{#1}}
\expandafter\ifx\csname urlstyle\endcsname\relax
  \providecommand{\doi}[1]{doi: #1}\else
  \providecommand{\doi}{doi: \begingroup \urlstyle{rm}\Url}\fi

\bibitem[Aukerman et~al.(2020)Aukerman, Carri{\`e}re, Chen, Gardner,
  Rabad{\'a}n, and Vanguri]{aukerman2020persistent}
Aukerman, A., Carri{\`e}re, M., Chen, C., Gardner, K., Rabad{\'a}n, R., and
  Vanguri, R.
\newblock Persistent homology based characterization of the breast cancer
  immune microenvironment: A feasibility study.
\newblock In \emph{36th International Symposium on Computational Geometry
  (SoCG)}, 2020.

\bibitem[Bhatia et~al.(2019)Bhatia, Chatterjee, Nathani, and
  Kaul]{bhatia2019persistent}
Bhatia, S., Chatterjee, B., Nathani, D., and Kaul, M.
\newblock A persistent homology perspective to the link prediction problem.
\newblock In \emph{International Conference on Complex Networks and Their
  Applications}, pp.\  27--39. Springer, 2019.

\bibitem[Bodnar et~al.(2021{\natexlab{a}})Bodnar, Frasca, Otter, Wang, Li{\`o},
  Mont{\'u}far, and Bronstein]{bodnar2021weisfeiler2}
Bodnar, C., Frasca, F., Otter, N., Wang, Y.~G., Li{\`o}, P., Mont{\'u}far, G.,
  and Bronstein, M.
\newblock Weisfeiler and lehman go cellular: Cw networks.
\newblock \emph{arXiv preprint arXiv:2106.12575}, 2021{\natexlab{a}}.

\bibitem[Bodnar et~al.(2021{\natexlab{b}})Bodnar, Frasca, Wang, Otter,
  Mont{\'u}far, Lio, and Bronstein]{bodnar2021weisfeiler1}
Bodnar, C., Frasca, F., Wang, Y.~G., Otter, N., Mont{\'u}far, G., Lio, P., and
  Bronstein, M.
\newblock Weisfeiler and lehman go topological: Message passing simplicial
  networks.
\newblock \emph{arXiv preprint arXiv:2103.03212}, 2021{\natexlab{b}}.

\bibitem[Bordes et~al.(2013)Bordes, Usunier, Garcia-Duran, Weston, and
  Yakhnenko]{bordes2013translating}
Bordes, A., Usunier, N., Garcia-Duran, A., Weston, J., and Yakhnenko, O.
\newblock Translating embeddings for modeling multi-relational data.
\newblock In \emph{Neural Information Processing Systems (NIPS)}, pp.\  1--9,
  2013.

\bibitem[Busaryev et~al.(2012)Busaryev, Cabello, Chen, Dey, and
  Wang]{busaryev2012annotating}
Busaryev, O., Cabello, S., Chen, C., Dey, T.~K., and Wang, Y.
\newblock Annotating simplices with a homology basis and its applications.
\newblock In \emph{Scandinavian workshop on algorithm theory}, pp.\  189--200.
  Springer, 2012.

\bibitem[Cai et~al.(2021)Cai, Li, Wang, and Ji]{cai2021line}
Cai, L., Li, J., Wang, J., and Ji, S.
\newblock Line graph neural networks for link prediction.
\newblock \emph{IEEE Transactions on Pattern Analysis and Machine
  Intelligence}, 2021.

\bibitem[Carri{\`e}re et~al.(2020)Carri{\`e}re, Chazal, Ike, Lacombe, Royer,
  and Umeda]{carriere2020perslay}
Carri{\`e}re, M., Chazal, F., Ike, Y., Lacombe, T., Royer, M., and Umeda, Y.
\newblock Perslay: A neural network layer for persistence diagrams and new
  graph topological signatures.
\newblock In \emph{International Conference on Artificial Intelligence and
  Statistics}, pp.\  2786--2796. PMLR, 2020.

\bibitem[Chambers et~al.(2009)Chambers, Erickson, and
  Nayyeri]{chambers2009minimum}
Chambers, E.~W., Erickson, J., and Nayyeri, A.
\newblock Minimum cuts and shortest homologous cycles.
\newblock In \emph{Proceedings of the twenty-fifth annual symposium on
  Computational geometry}, pp.\  377--385, 2009.

\bibitem[Chan et~al.(2013)Chan, Carlsson, and Rabadan]{chan2013topology}
Chan, J.~M., Carlsson, G., and Rabadan, R.
\newblock Topology of viral evolution.
\newblock \emph{Proceedings of the National Academy of Sciences}, 110\penalty0
  (46):\penalty0 18566--18571, 2013.

\bibitem[Chen \& Freedman(2010)Chen and Freedman]{chen2010measuring}
Chen, C. and Freedman, D.
\newblock Measuring and computing natural generators for homology groups.
\newblock \emph{Computational Geometry}, 43\penalty0 (2):\penalty0 169--181,
  2010.

\bibitem[Chen \& Freedman(2011)Chen and Freedman]{chen2011hardness}
Chen, C. and Freedman, D.
\newblock Hardness results for homology localization.
\newblock \emph{Discrete \& Computational Geometry}, 45\penalty0 (3):\penalty0
  425--448, 2011.

\bibitem[Chen et~al.(2019)Chen, Ni, Bai, and Wang]{chen2019topological}
Chen, C., Ni, X., Bai, Q., and Wang, Y.
\newblock A topological regularizer for classifiers via persistent homology.
\newblock In \emph{The 22nd International Conference on Artificial Intelligence
  and Statistics}, pp.\  2573--2582. PMLR, 2019.

\bibitem[Chen et~al.(2020)Chen, Wang, Zhao, Cheng, Zhao, and
  Duan]{chen2020knowledge}
Chen, Z., Wang, Y., Zhao, B., Cheng, J., Zhao, X., and Duan, Z.
\newblock Knowledge graph completion: A review.
\newblock \emph{IEEE Access}, 8:\penalty0 192435--192456, 2020.
\newblock \doi{10.1109/ACCESS.2020.3030076}.

\bibitem[Dettmers et~al.(2018)Dettmers, Minervini, Stenetorp, and
  Riedel]{dettmers2018convolutional}
Dettmers, T., Minervini, P., Stenetorp, P., and Riedel, S.
\newblock Convolutional 2d knowledge graph embeddings.
\newblock In \emph{Thirty-second AAAI conference on artificial intelligence},
  2018.

\bibitem[Dey \& Wang(2022)Dey and Wang]{dey2022computational}
Dey, T.~K. and Wang, Y.
\newblock \emph{Computational Topology for Data Analysis}.
\newblock Cambridge University Press, 2022.

\bibitem[Dey et~al.(2010)Dey, Sun, and Wang]{dey2010approximating}
Dey, T.~K., Sun, J., and Wang, Y.
\newblock Approximating loops in a shortest homology basis from point data.
\newblock In \emph{Proceedings of the twenty-sixth annual symposium on
  Computational geometry}, pp.\  166--175, 2010.

\bibitem[Dey et~al.(2011)Dey, Hirani, and Krishnamoorthy]{dey2011optimal}
Dey, T.~K., Hirani, A.~N., and Krishnamoorthy, B.
\newblock Optimal homologous cycles, total unimodularity, and linear
  programming.
\newblock \emph{SIAM Journal on Computing}, 40\penalty0 (4):\penalty0
  1026--1044, 2011.

\bibitem[Dey et~al.(2020)Dey, Hou, and Mandal]{dey2020computing}
Dey, T.~K., Hou, T., and Mandal, S.
\newblock Computing minimal persistent cycles: Polynomial and hard cases.
\newblock In \emph{Proceedings of the Fourteenth Annual ACM-SIAM Symposium on
  Discrete Algorithms}, pp.\  2587--2606. SIAM, 2020.

\bibitem[Edelsbrunner \& Harer(2010)Edelsbrunner and
  Harer]{edelsbrunner2010computational}
Edelsbrunner, H. and Harer, J.
\newblock \emph{Computational topology: an introduction}.
\newblock American Mathematical Soc., 2010.

\bibitem[Edelsbrunner et~al.(2000)Edelsbrunner, Letscher, and
  Zomorodian]{edelsbrunner2000topological}
Edelsbrunner, H., Letscher, D., and Zomorodian, A.
\newblock Topological persistence and simplification.
\newblock In \emph{Proceedings 41st annual symposium on foundations of computer
  science}, pp.\  454--463. IEEE, 2000.

\bibitem[Gal{\'a}rraga et~al.(2013)Gal{\'a}rraga, Teflioudi, Hose, and
  Suchanek]{galarraga2013amie}
Gal{\'a}rraga, L.~A., Teflioudi, C., Hose, K., and Suchanek, F.
\newblock Amie: association rule mining under incomplete evidence in
  ontological knowledge bases.
\newblock In \emph{Proceedings of the 22nd international conference on World
  Wide Web}, pp.\  413--422, 2013.

\bibitem[Gao \& Ji(2019)Gao and Ji]{gao2019graph}
Gao, H. and Ji, S.
\newblock Graph u-nets.
\newblock In Chaudhuri, K. and Salakhutdinov, R. (eds.), \emph{Proceedings of
  the 36th International Conference on Machine Learning, {ICML} 2019, 9-15 June
  2019, Long Beach, California, {USA}}, volume~97 of \emph{Proceedings of
  Machine Learning Research}, pp.\  2083--2092. {PMLR}, 2019.
\newblock URL \url{http://proceedings.mlr.press/v97/gao19a.html}.

\bibitem[Giusti et~al.(2015)Giusti, Pastalkova, Curto, and
  Itskov]{giusti2015clique}
Giusti, C., Pastalkova, E., Curto, C., and Itskov, V.
\newblock Clique topology reveals intrinsic geometric structure in neural
  correlations.
\newblock \emph{Proceedings of the National Academy of Sciences}, 112\penalty0
  (44):\penalty0 13455--13460, 2015.

\bibitem[Hofer et~al.(2017)Hofer, Kwitt, Niethammer, and Uhl]{hofer2017deep}
Hofer, C., Kwitt, R., Niethammer, M., and Uhl, A.
\newblock Deep learning with topological signatures.
\newblock In \emph{Proceedings of the 31st International Conference on Neural
  Information Processing Systems}, pp.\  1633--1643, 2017.

\bibitem[Hofer et~al.(2020)Hofer, Graf, Rieck, Niethammer, and
  Kwitt]{hofer2020graph}
Hofer, C., Graf, F., Rieck, B., Niethammer, M., and Kwitt, R.
\newblock Graph filtration learning.
\newblock In \emph{International Conference on Machine Learning}, pp.\
  4314--4323. PMLR, 2020.

\bibitem[Hu et~al.(2019)Hu, Li, Samaras, and Chen]{hu2019topology}
Hu, X., Li, F., Samaras, D., and Chen, C.
\newblock Topology-preserving deep image segmentation.
\newblock \emph{Advances in Neural Information Processing Systems}, 32, 2019.

\bibitem[Hu et~al.(2021)Hu, Wang, Fuxin, Samaras, and Chen]{hu2021topology}
Hu, X., Wang, Y., Fuxin, L., Samaras, D., and Chen, C.
\newblock Topology-aware segmentation using discrete morse theory.
\newblock In \emph{The Ninth International Conference on Learning
  Representations (ICLR)}, 2021.

\bibitem[Huang et~al.(2019)Huang, Zhang, Li, and Li]{huang2019knowledge}
Huang, X., Zhang, J., Li, D., and Li, P.
\newblock Knowledge graph embedding based question answering.
\newblock In \emph{Proceedings of the Twelfth ACM International Conference on
  Web Search and Data Mining}, pp.\  105--113, 2019.

\bibitem[Kampffmeyer et~al.(2019)Kampffmeyer, Chen, Liang, Wang, Zhang, and
  Xing]{kampffmeyer2019rethinking}
Kampffmeyer, M., Chen, Y., Liang, X., Wang, H., Zhang, Y., and Xing, E.~P.
\newblock Rethinking knowledge graph propagation for zero-shot learning.
\newblock In \emph{Proceedings of the IEEE/CVF Conference on Computer Vision
  and Pattern Recognition}, pp.\  11487--11496, 2019.

\bibitem[Kanari et~al.(2018)Kanari, D{\l}otko, Scolamiero, Levi, Shillcock,
  Hess, and Markram]{kanari2018topological}
Kanari, L., D{\l}otko, P., Scolamiero, M., Levi, R., Shillcock, J., Hess, K.,
  and Markram, H.
\newblock A topological representation of branching neuronal morphologies.
\newblock \emph{Neuroinformatics}, 16\penalty0 (1):\penalty0 3--13, 2018.

\bibitem[Kipf \& Welling(2017)Kipf and Welling]{kipf2016semi}
Kipf, T.~N. and Welling, M.
\newblock Semi-supervised classification with graph convolutional networks.
\newblock In \emph{5th International Conference on Learning Representations,
  {ICLR} 2017, Toulon, France, April 24-26, 2017, Conference Track
  Proceedings}. OpenReview.net, 2017.
\newblock URL \url{https://openreview.net/forum?id=SJU4ayYgl}.

\bibitem[Langley(2000)]{langley00}
Langley, P.
\newblock Crafting papers on machine learning.
\newblock In Langley, P. (ed.), \emph{Proceedings of the 17th International
  Conference on Machine Learning (ICML 2000)}, pp.\  1207--1216, Stanford, CA,
  2000. Morgan Kaufmann.

\bibitem[Li et~al.(2017)Li, Wang, Ascoli, Mitra, and Wang]{li2017metrics}
Li, Y., Wang, D., Ascoli, G.~A., Mitra, P., and Wang, Y.
\newblock Metrics for comparing neuronal tree shapes based on persistent
  homology.
\newblock \emph{PloS one}, 12\penalty0 (8):\penalty0 e0182184, 2017.

\bibitem[Mai et~al.(2021)Mai, Zheng, Yang, and Hu]{mai2020communicative}
Mai, S., Zheng, S., Yang, Y., and Hu, H.
\newblock Communicative message passing for inductive relation reasoning.
\newblock In \emph{Proceedings of the AAAI Conference on Artificial
  Intelligence}, volume~35, pp.\  4294--4302, 2021.

\bibitem[Marcheggiani \& Titov(2017)Marcheggiani and
  Titov]{marcheggiani2017encoding}
Marcheggiani, D. and Titov, I.
\newblock Encoding sentences with graph convolutional networks for semantic
  role labeling.
\newblock In \emph{Proceedings of the 2017 Conference on Empirical Methods in
  Natural Language Processing}, pp.\  1506--1515, 2017.

\bibitem[Meilicke et~al.(2018)Meilicke, Fink, Wang, Ruffinelli, Gemulla, and
  Stuckenschmidt]{meilicke2018fine}
Meilicke, C., Fink, M., Wang, Y., Ruffinelli, D., Gemulla, R., and
  Stuckenschmidt, H.
\newblock Fine-grained evaluation of rule-and embedding-based systems for
  knowledge graph completion.
\newblock In \emph{International Semantic Web Conference}, pp.\  3--20.
  Springer, 2018.

\bibitem[Munkres(2018)]{munkres2018elements}
Munkres, J.~R.
\newblock \emph{Elements of algebraic topology}.
\newblock CRC press, 2018.

\bibitem[Petri et~al.(2014)Petri, Expert, Turkheimer, Carhart-Harris, Nutt,
  Hellyer, and Vaccarino]{petri2014homological}
Petri, G., Expert, P., Turkheimer, F., Carhart-Harris, R., Nutt, D., Hellyer,
  P.~J., and Vaccarino, F.
\newblock Homological scaffolds of brain functional networks.
\newblock \emph{Journal of The Royal Society Interface}, 11\penalty0
  (101):\penalty0 20140873, 2014.

\bibitem[Rizvi et~al.(2017)Rizvi, Camara, Kandror, Roberts, Schieren, Maniatis,
  and Rabadan]{rizvi2017single}
Rizvi, A.~H., Camara, P.~G., Kandror, E.~K., Roberts, T.~J., Schieren, I.,
  Maniatis, T., and Rabadan, R.
\newblock Single-cell topological rna-seq analysis reveals insights into
  cellular differentiation and development.
\newblock \emph{Nature biotechnology}, 35\penalty0 (6):\penalty0 551--560,
  2017.

\bibitem[Sadeghian et~al.(2019)Sadeghian, Armandpour, Ding, and
  Wang]{sadeghian2019drum}
Sadeghian, A., Armandpour, M., Ding, P., and Wang, D.~Z.
\newblock Drum: End-to-end differentiable rule mining on knowledge graphs.
\newblock \emph{arXiv preprint arXiv:1911.00055}, 2019.

\bibitem[Sun et~al.(2019)Sun, Vashishth, Sanyal, Talukdar, and Yang]{sun2019re}
Sun, Z., Vashishth, S., Sanyal, S., Talukdar, P., and Yang, Y.
\newblock A re-evaluation of knowledge graph completion methods.
\newblock \emph{arXiv preprint arXiv:1911.03903}, 2019.

\bibitem[Teru et~al.(2020)Teru, Denis, and Hamilton]{teru2020inductive}
Teru, K., Denis, E., and Hamilton, W.
\newblock Inductive relation prediction by subgraph reasoning.
\newblock In \emph{International Conference on Machine Learning}, pp.\
  9448--9457. PMLR, 2020.

\bibitem[Toutanova \& Chen(2015)Toutanova and Chen]{toutanova2015observed}
Toutanova, K. and Chen, D.
\newblock Observed versus latent features for knowledge base and text
  inference.
\newblock In \emph{Proceedings of the 3rd workshop on continuous vector space
  models and their compositionality}, pp.\  57--66, 2015.

\bibitem[Vashishth et~al.(2019)Vashishth, Sanyal, Nitin, and
  Talukdar]{vashishth2019composition}
Vashishth, S., Sanyal, S., Nitin, V., and Talukdar, P.
\newblock Composition-based multi-relational graph convolutional networks.
\newblock In \emph{International Conference on Learning Representations}, 2019.

\bibitem[Wang et~al.(2020)Wang, Liu, Samaras, and Chen]{wang2020topogan}
Wang, F., Liu, H., Samaras, D., and Chen, C.
\newblock Topogan: A topology-aware generative adversarial network.
\newblock In \emph{European Conference on Computer Vision}, pp.\  118--136.
  Springer, 2020.

\bibitem[Wang et~al.(2021)Wang, Kapse, Liu, Prasanna, and
  Chen]{wang2021topotxr}
Wang, F., Kapse, S., Liu, S., Prasanna, P., and Chen, C.
\newblock Topotxr: A topological biomarker for predicting treatment response in
  breast cancer.
\newblock In \emph{International Conference on Information Processing in
  Medical Imaging}, pp.\  386--397. Springer, 2021.

\bibitem[Wang et~al.(2018)Wang, Zhang, Wang, Zhao, Li, Xie, and
  Guo]{wang2018ripplenet}
Wang, H., Zhang, F., Wang, J., Zhao, M., Li, W., Xie, X., and Guo, M.
\newblock Ripplenet: Propagating user preferences on the knowledge graph for
  recommender systems.
\newblock In \emph{Proceedings of the 27th ACM International Conference on
  Information and Knowledge Management}, pp.\  417--426, 2018.

\bibitem[Wu et~al.(2017)Wu, Chen, Wang, Zhang, Yuan, Qian, Metaxas, and
  Axel]{wu2017optimal}
Wu, P., Chen, C., Wang, Y., Zhang, S., Yuan, C., Qian, Z., Metaxas, D., and
  Axel, L.
\newblock Optimal topological cycles and their application in cardiac
  trabeculae restoration.
\newblock In \emph{International Conference on Information Processing in
  Medical Imaging}, pp.\  80--92. Springer, 2017.

\bibitem[Wu et~al.(2020)Wu, Zheng, Goswami, Metaxas, and
  Chen]{wu2020topological}
Wu, P., Zheng, S., Goswami, M., Metaxas, D., and Chen, C.
\newblock A topological filter for learning with label noise.
\newblock \emph{Advances in neural information processing systems},
  33:\penalty0 21382--21393, 2020.

\bibitem[Xiong et~al.(2017)Xiong, Hoang, and Wang]{xiong2017deeppath}
Xiong, W., Hoang, T., and Wang, W.~Y.
\newblock Deeppath: A reinforcement learning method for knowledge graph
  reasoning.
\newblock \emph{arXiv preprint arXiv:1707.06690}, 2017.

\bibitem[Xu et~al.(2019)Xu, Hu, Leskovec, and Jegelka]{xu2019powerful}
Xu, K., Hu, W., Leskovec, J., and Jegelka, S.
\newblock How powerful are graph neural networks?
\newblock In \emph{7th International Conference on Learning Representations,
  {ICLR} 2019, New Orleans, LA, USA, May 6-9, 2019}. OpenReview.net, 2019.
\newblock URL \url{https://openreview.net/forum?id=ryGs6iA5Km}.

\bibitem[Yan et~al.(2021)Yan, Ma, Gao, Tang, and Chen]{yan2021link}
Yan, Z., Ma, T., Gao, L., Tang, Z., and Chen, C.
\newblock Link prediction with persistent homology: An interactive view.
\newblock In \emph{International Conference on Machine Learning}, pp.\
  11659--11669. PMLR, 2021.

\bibitem[Yan et~al.(2022)Yan, Ma, Gao, Tang, Wang, and Chen]{yan2022neural}
Yan, Z., Ma, T., Gao, L., Tang, Z., Wang, Y., and Chen, C.
\newblock Neural approximation of extended persistent homology on graphs.
\newblock In \emph{ICLR 2022 Workshop on Geometrical and Topological
  Representation Learning}, 2022.

\bibitem[Yang et~al.(2014)Yang, Yih, He, Gao, and Deng]{yang2014embedding}
Yang, B., Yih, W.-t., He, X., Gao, J., and Deng, L.
\newblock Embedding entities and relations for learning and inference in
  knowledge bases.
\newblock \emph{arXiv preprint arXiv:1412.6575}, 2014.

\bibitem[Yang et~al.(2017)Yang, Yang, and Cohen]{yang2017differentiable}
Yang, F., Yang, Z., and Cohen, W.~W.
\newblock Differentiable learning of logical rules for knowledge base
  reasoning.
\newblock \emph{arXiv preprint arXiv:1702.08367}, 2017.

\bibitem[Ye et~al.(2019)Ye, Liu, Ma, Gao, and Chen]{ye2019curvature}
Ye, Z., Liu, K.~S., Ma, T., Gao, J., and Chen, C.
\newblock Curvature graph network.
\newblock In \emph{International Conference on Learning Representations}, 2019.

\bibitem[You et~al.(2019)You, Ying, and Leskovec]{you2019position}
You, J., Ying, R., and Leskovec, J.
\newblock Position-aware graph neural networks.
\newblock In \emph{International Conference on Machine Learning}, pp.\
  7134--7143. PMLR, 2019.

\bibitem[Zhang et~al.(2018)Zhang, Dai, Kozareva, Smola, and
  Song]{zhang2018variational}
Zhang, Y., Dai, H., Kozareva, Z., Smola, A., and Song, L.
\newblock Variational reasoning for question answering with knowledge graph.
\newblock In \emph{Proceedings of the AAAI Conference on Artificial
  Intelligence}, volume~32, 2018.

\bibitem[Zhao \& Wang(2019)Zhao and Wang]{zhao2019learning}
Zhao, Q. and Wang, Y.
\newblock Learning metrics for persistence-based summaries and applications for
  graph classification.
\newblock \emph{Advances in Neural Information Processing Systems},
  32:\penalty0 9859--9870, 2019.

\bibitem[Zhao et~al.(2020{\natexlab{a}})Zhao, Ye, Chen, and
  Wang]{zhao2020persistence}
Zhao, Q., Ye, Z., Chen, C., and Wang, Y.
\newblock Persistence enhanced graph neural network.
\newblock In \emph{International Conference on Artificial Intelligence and
  Statistics}, pp.\  2896--2906. PMLR, 2020{\natexlab{a}}.

\bibitem[Zhao et~al.(2020{\natexlab{b}})Zhao, Qin, Liu, and
  Wang]{zhao2020biomedical}
Zhao, S., Qin, B., Liu, T., and Wang, F.
\newblock Biomedical knowledge graph refinement with embedding and logic rules.
\newblock \emph{arXiv preprint arXiv:2012.01031}, 2020{\natexlab{b}}.

\bibitem[Zheng et~al.(2021)Zheng, Zhang, Wagner, Goswami, and
  Chen]{zheng2021topological}
Zheng, S., Zhang, Y., Wagner, H., Goswami, M., and Chen, C.
\newblock Topological detection of trojaned neural networks.
\newblock \emph{Advances in Neural Information Processing Systems}, 34, 2021.

\bibitem[Zhu et~al.(2020)Zhu, Che, Jin, Zhang, Su, and Wang]{zhu2020knowledge}
Zhu, Y., Che, C., Jin, B., Zhang, N., Su, C., and Wang, F.
\newblock Knowledge-driven drug repurposing using a comprehensive drug
  knowledge graph.
\newblock \emph{Health Informatics Journal}, 26\penalty0 (4):\penalty0
  2737--2750, 2020.

\bibitem[Zhu et~al.(2021)Zhu, Zhang, Xhonneux, and Tang]{zhu2021neural}
Zhu, Z., Zhang, Z., Xhonneux, L.-P., and Tang, J.
\newblock Neural bellman-ford networks: A general graph neural network
  framework for link prediction.
\newblock \emph{Advances in Neural Information Processing Systems}, 34, 2021.

\end{thebibliography}
\bibliographystyle{icml2022}

\newpage
\appendix
\onecolumn
\section{Appendix}
\subsection{BR-LSTM encoding}

\begin{figure}[hbtp]
	\centering
	\vspace{-.1in}
	\includegraphics[width=0.7\textwidth]{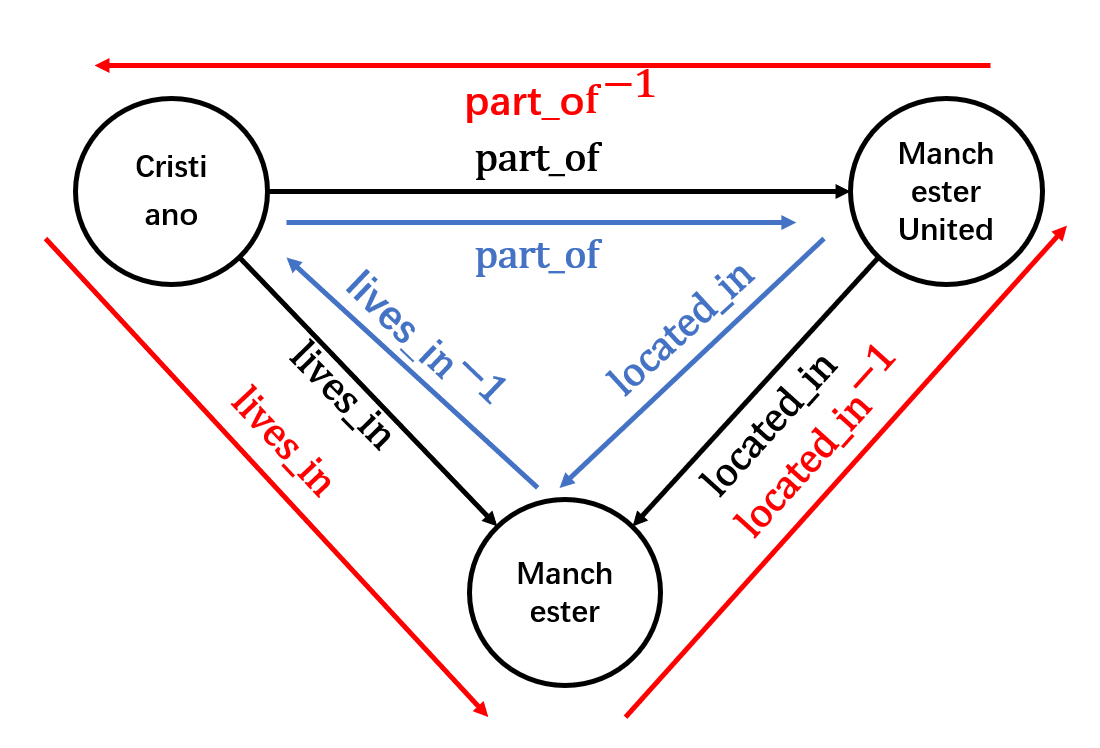}
	\caption{An example of BR-LSTM construction for Figure~\ref{fig:motiv} (b).}
	\label{fig:bi}
	\vspace{-.1in}
\end{figure}

\textbf{An example of BR-LSTM construction.} In this section, we provide an example of the construction of BR-LSTM proposed in Section~\ref{subsec:tree}. Recall that in the generation of cycle feature, we tackle the input KG as a directed graph with different edges/triplets associated with different relations. We assume that information in an edge/triplet flows along both directions, and encode the cycle in a relation-aware manner. 
We denote by $(u, r, v)\in E$ a triplet connecting nodes $u$ and $v$ with relation $r$. Here $E$ is the set of all the triplets, we assume that an inverse triplet $(v, r^{-1}, u)$ is also included in the KG. Here $r^{-1}$ is defined as the inverse relation of $r$. 
Formally, we extend the triplet set of the KG as: $E' = E \cup \{(v,r^{-1},u)| (u,r,v) \in E\}$. An illustration is shown in Figure~\ref{fig:bi}. Through the triplets in $E'$, we can convert the rule shown in Figure~\ref{fig:motiv} (b) into two opposite sequences: $(Cristiano, part\_of, Manchester United)$, $(Manchester United, located\_in, Manchester)$, $(Manchester, lives\_in^{-1}, Cristiano)$ and $(Manchester United, part\_of^{-1}, Cristiano)$, $(Cristiano, lives\_in, Manchester)$, $(Manchester, located\_in^{-1}, Manchester United)$.

\textbf{The sequence of relations in the cycles.} In the encoding of cycles, there can be several different sequences of rules. In our setting, we choose the target triplet (the non-tree edge in the cycle) as the start token of BR-LSTM. Our aim is to preserve the sequence of the rule, and let the model focus on predicting the target triplet. An examples is shown in Figure~\ref{fig:bi}: the rule $\exists X, (X, \text{ \textit{part\_of} }, Y) \wedge (X, \text{ \textit{lives\_in} }, Z) \rightarrow (Y, \text{ \textit{located\_in} }, Z)$ and the other rule $\exists Z, (X, \text{ \textit{lives\_in} }, Z) \wedge (Y, \text{ \textit{located\_in} }, Z) \rightarrow (X, \text{ \textit{part\_of} }, Y)$ should have different confidence values. The former one is a good rule because if a player $X$ is a part of team $Y$ and $X$ also lives in the city $Z$, then the team $Y$ should be also located in city $Z$. While in the later rule, if a player $X$ lives in city $Z$, and a team $Y$ also located in city $Z$, $X$ does not necessarily need to play for team $Y$.

\begin{table}[btp]
	\caption{Statistics of inductive benchmarks.}
	\label{tab:data_ind}
	\vskip 0.15in
	\begin{center}
		\begin{small}
			\begin{sc}
				\scalebox{0.75}{
					\begin{tabular}{|lc|ccc|ccc|ccc|}
						\hline\noalign{\smallskip}
						& & \multicolumn{3}{c|}{WN18RR} &\multicolumn{3}{c|}{FB15K-237}&\multicolumn{3}{c|}{NELL-995} \\
						\noalign{\smallskip}\hline\noalign{\smallskip}
						&  & relations & nodes & links & relations &  nodes & links & relations & nodse & links\\
						\noalign{\smallskip}\hline\noalign{\smallskip}
						v1  & train  & 9 & 2746 & 6678 & 183 & 2000 & 5226 & 14 & 10915 & 5540 \\
						& test & 9 & 922 & 1991 & 146 & 1500 & 2404 & 14 & 225 & 1034  \\
						\noalign{\smallskip}\hline\noalign{\smallskip}
						v2   & train & 10 & 6954 & 18968 & 203 & 3000 & 12085 & 88 & 2564 & 10109 \\
						& test & 10 & 2923 & 4863 & 176 & 2000 & 5092 & 79 & 4937 & 5521 \\
						\noalign{\smallskip}\hline\noalign{\smallskip} 
						v3 & train & 11 & 12078 & 32150 & 218 & 4000 & 22394 & 142 & 4647 & 20117  \\
						& test & 11 & 5084 & 7470 & 187 & 3000 & 9137 & 122 & 4921 & 9668  \\
						\noalign{\smallskip}\hline\noalign{\smallskip} 
						v4 & train & 9  & 3861 & 9842 & 222 & 5000 & 33916 & 77 & 2092 & 9289\\
						& test & 9 & 7208 & 15157 & 204 & 3500 & 14554 & 61 & 3294 & 8520\\
						\noalign{\smallskip}\hline
				\end{tabular}}
			\end{sc}
		\end{small}
	\end{center}
	\vskip -0.1in
\end{table}

\subsection{Experimental details.}

\label{subsec:details}

\textbf{Datasets.} The datasets used in our settings are subsets of KG WN18RR~\citep{toutanova2015observed}, FB15k-237~\citep{dettmers2018convolutional}, and NELL-995~\citep{xiong2017deeppath}. \citet{teru2020inductive} generate these datasets by sampling disjoint subgraphs from the original datasets. For inductive relation prediction, the entities in the training set and the test set should not be overlapped. To evaluate the robustness of models, \citet{teru2020inductive} sample four different pairs of training sets and test sets with the increasing number of nodes and edges. The details of the benchmark datasets are shown in Table~\ref{tab:data_ind}.

\textbf{Experimental details.}  We adopt a 2-layer BR-LSTM to generate feature vectors for all the cycles in a cycle basis. Its output feature vector dimension is set to 20. A 2-layer GCN~\citep{kipf2016semi} is adopted for the message passing of cycle basis, where ReLU serves as the activation function between GCN layers. We combine 20 different shortest path trees to learn the good rules in the given dataset\footnote{In seldom cases such as NELL-995 v2, considering that we can significantly benefit from more shortest path trees, we combine 50 cycle bases for relation prediction.}. In the cycle graph, we select the top 2 most related cycles for each cycle. For all the modules, Adam is used as the optimizer, the dropout is set to 0.2, the epoch is set to 100 with an early-stopping of 20, the learning rate is 0.005 and the weight decay is 5e-5. We follow the settings in ~\citep{teru2020inductive, mai2020communicative}, that is, to view all the existing triplets in KG as positive triplets and sample negative triplets by replacing the head (or tail) of the triplet with a uniformly sampled random entity. We use binary cross-entropy loss as the loss function with the negative sampling method. Considering that some inductive test sets contain few cycles, which leads to the inconsistent performance between the inductive test sets and original training sets, we use the inductive training set as the validation set (while the training set and the test set are the same with \citep{teru2020inductive, mai2020communicative}). We run all the baseline methods with a cluster of two Intel Xeon Gold 5128 processors, 192GB RAM, and one GeForce RTX 2080 Ti graphics card.

\subsection{Additional Experiments}



\begin{center}
	\begin{table}[btp]
		\centering
		\caption{AUC-PR scores of inductive relation prediction, we keep the validation datasets of baseline methods as the same as ours and run these methods five times for the average scores.}
		\label{tab:new} 
		\scalebox{0.71}{
			\begin{tabular}{|l|cccc|cccc|cccc|}
				\hline\noalign{\smallskip}
				&\multicolumn{4}{c|}{WN18RR} &\multicolumn{4}{c|}{FB15K-237}&\multicolumn{4}{c|}{NELL-995} \\
				\noalign{\smallskip}\hline\noalign{\smallskip}
				Method & v1 & v2 & v3 & v4 & v1 & v2  & v3 & v4&  v1 & v2 & v3 & v4\\
				\noalign{\smallskip}\hline\noalign{\smallskip}
				GraIL & 96.09 & 95.92  & 85.86 & 94.02 & 
				86.75 & 91.84 & 90.17 & 84.74 & 82.34 & 92.35 & 91.45 & 82.88 \\
				CoMPILE & 98.56 & \textbf{99.98} & 94.04  & \textbf{99.85} & 83.45 &  92.17& 90.91  &  91.39 &  78.07 & 94.07  & 95.69 & 83.40\\
				NBFNet & 97.87 & 97.48 & \textbf{95.67} & 96.67 & 92.81 & 96.40 & 95.79 & 94.67 & \textbf{88.82} & 93.78 & 95.32 & 90.10\\
				\noalign{\smallskip}\hline\noalign{\smallskip}
				CBGNN & \textbf{98.63} & 97.62 & 89.76 & 97.80 & \textbf{96.34} & \textbf{96.53} & \textbf{96.38} & \textbf{95.23} & 82.79
				& \textbf{94.78} & \textbf{96.29}& \textbf{94.02}\\
				\noalign{\smallskip}\hline
		\end{tabular}}
	\end{table}
\end{center}
\begin{center}
	\begin{table}[btp]
		\centering
		\caption{Hit@10 scores of inductive relation prediction, we keep the validation datasets of baseline methods as the same as ours and run these methods five times for the average scores.}
		\label{tab:new_hits} 
		\scalebox{0.75}{
			\begin{tabular}{|l|cccc|cccc|cccc|}
				\hline\noalign{\smallskip}
				&\multicolumn{4}{c|}{WN18RR} &\multicolumn{4}{c|}{FB15K-237}&\multicolumn{4}{c|}{NELL-995} \\
				\noalign{\smallskip}\hline\noalign{\smallskip}
				Method & v1 & v2 & v3 & v4 & v1 & v2  & v3 & v4&  v1 & v2 & v3 & v4\\
				\noalign{\smallskip}\hline\noalign{\smallskip}
				GraIL & 84.04 & 81.63 & 60.65 & 75.34 & 66.30
				& 82.00 & 82.54 & 78.16 & 55.00 & 93.27 & 89.74 & 73.94\\
				CoMPILE & 82.71 & 80.82 & 62.56 & 75.92 & 69.75  & 82.52 & 82.95 & 85.46 & 62.00 & 91.18 & 93.75 & 74.29\\
				NBFNet & 93.14 & 90.56 & \textbf{90.09} & 88.58 & 82.11 & 94.95 & 94.72 & 94.20 & 62.71 & 89.56 & 95.10 & 82.12\\
				\noalign{\smallskip}\hline\noalign{\smallskip}
				CBGNN & \textbf{98.40} & \textbf{96.14}  & 62.28 &\textbf{96.50} & \textbf{97.56}  & \textbf{96.03} & \textbf{94.91} & \textbf{94.73} &  \textbf{84.00}
				& \textbf{94.96} & \textbf{95.34} & \textbf{92.34}\\
				\noalign{\smallskip}\hline
		\end{tabular}}
	\end{table}
\end{center}

\textbf{Experiments with the same settings.} In Section~\ref{sec:exp}, we have compared the performance of our model with the baseline results copied from \citep{teru2020inductive, mai2020communicative}, as shown in Table~\ref{tab:expres} and Table~\ref{tab:hits}. For a fair comparison, we set Grail and CoMPILE as the same experimental settings (the same validation setting, as stated in Experimental details) as ours, and record the result in Table~\ref{tab:new} and Table~\ref{tab:new_hits}. We also note that a relevant work NBFNet~\cite{zhu2021neural} is published recently. Considering that they do not provide the the results on all datasets in the original paper, we add the comparison here. Similar to the observation in Section~\ref{sec:exp}, CBGNN consistently achieves the state-of-the-art results in the evaluation of Hit@10 scores and outperforms the majority of benchmark datasets when it comes to AUC-PR scores. The results further show the effectiveness of our proposed method.

\begin{figure}[btp]
	\centering
	\subfigure[]{
		\begin{minipage}[t]{0.5\linewidth}
			\centering
			\includegraphics[width=\columnwidth]{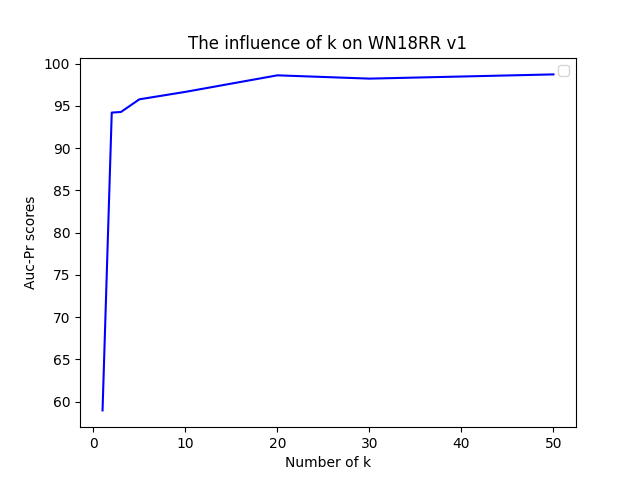}
		\end{minipage}
	}%
	\subfigure[]{
		\begin{minipage}[t]{0.5\linewidth}
			\centering
			\includegraphics[width=\columnwidth]{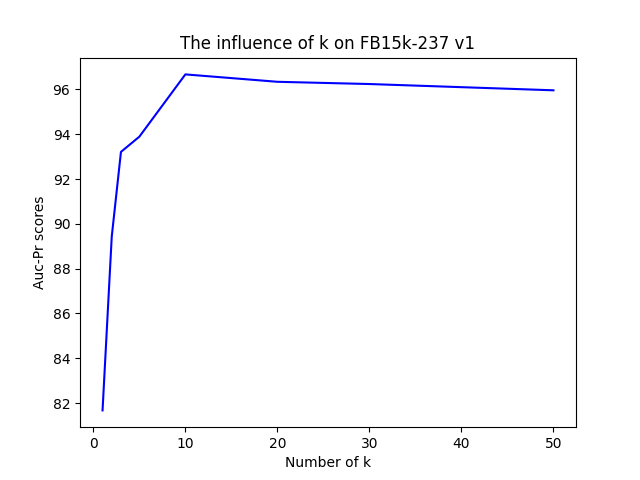}
		\end{minipage}%
	}\\%
	\subfigure[]{
		\begin{minipage}[t]{0.5\linewidth}
			\centering
			\includegraphics[width=\columnwidth]{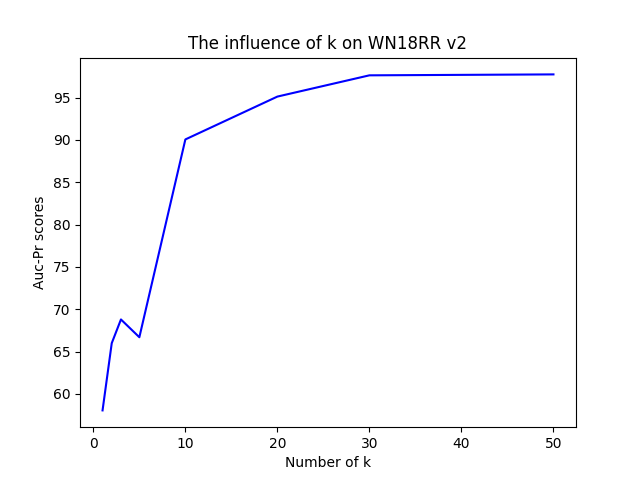}
		\end{minipage}
	}%
	\subfigure[]{
		\begin{minipage}[t]{0.5\linewidth}
			\centering
			\includegraphics[width=\columnwidth]{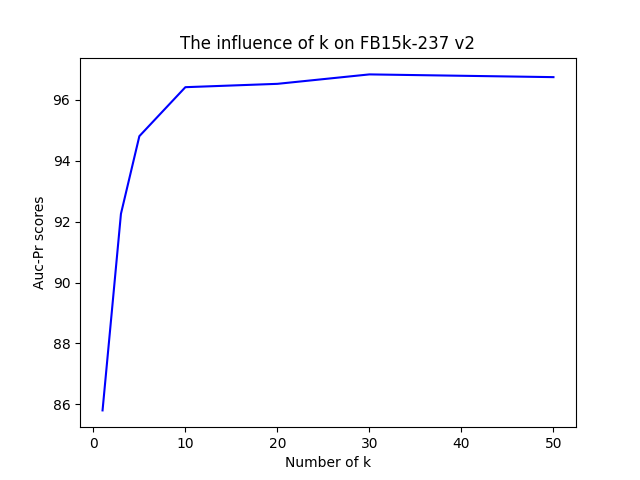}
		\end{minipage}%
	}\\%
	\centering
	\vspace{-.1in}
	\caption{The influence of $k$ on WN18RR v1, v2, and FB15k-237 v1, v2.}
	\label{fig:k}
	\vspace{-.15in}
\end{figure}

\textbf{The influence of $k$.} 
In this paragraph, we do experiments on the influence of the number of the shortest path trees $k$ which are used to learn the suitable cycle basis. As is shown in Figure~\ref{fig:k}, CBGNN performs badly with a single cycle basis. However, its performance grows quickly as $k$ increases from 1, and gradually converges after $k$ is large enough (10 for smaller graphs like WN18RR v1 and FB15k-237 v1, and 20 for larger graphs like WN18RR v2 and FB15k-237 v2). The experiments show that it is crucial to utilize multiple bases to guarantee better coverage. However, after $k$ grows to a certain extent, the root nodes will be spread out, and contain enough information to cover the whole graph. Therefore, the model hardly benefits from the increase of $k$ after it is larger than a certain threshold. One important factor that may influence the threshold is the size of the input graph. For smaller graphs, we only need a small number of SPT cycle bases to cover of the graph. While for larger graphs, we may need more SPT cycle bases. But as shown in Table \ref{tab:expres} and \ref{tab:hits}, 20 SPT cycle bases are enough to gain a state-of-the-art results in most situations.

\begin{figure}[btp]
	\centering
	\subfigure[]{
		\begin{minipage}[t]{0.5\linewidth}
			\centering
			\includegraphics[width=\columnwidth]{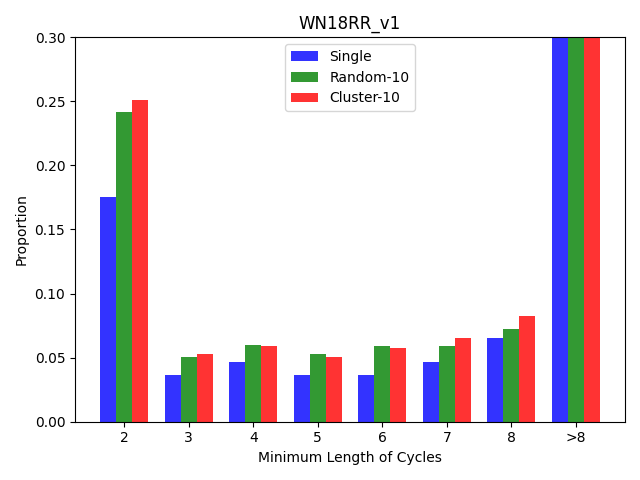}
		\end{minipage}
	}%
	\subfigure[]{
		\begin{minipage}[t]{0.5\linewidth}
			\centering
			\includegraphics[width=\columnwidth]{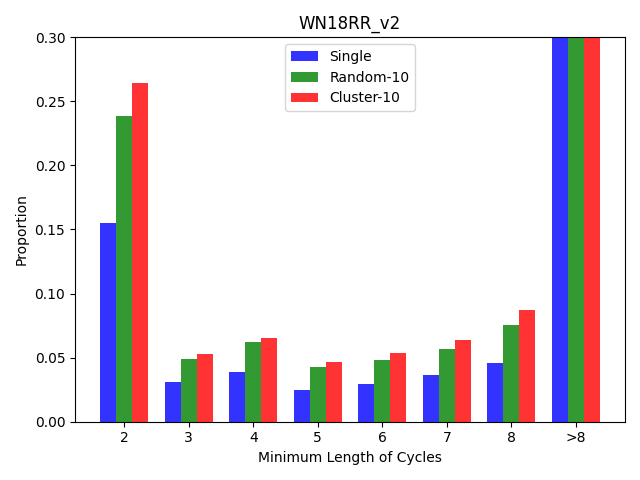}
		\end{minipage}%
	}%
	\\
	\subfigure[]{
		\begin{minipage}[t]{0.5\linewidth}
			\centering
			\includegraphics[width=\columnwidth]{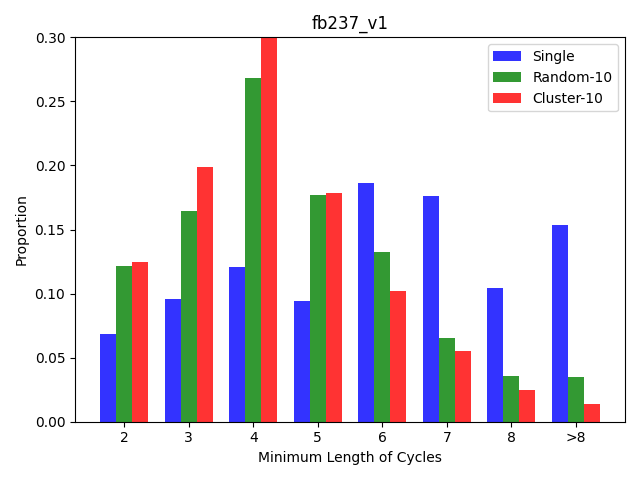}
		\end{minipage}%
	}%
	\subfigure[]{
		\begin{minipage}[t]{0.5\linewidth}
			\centering
			\includegraphics[width=\columnwidth]{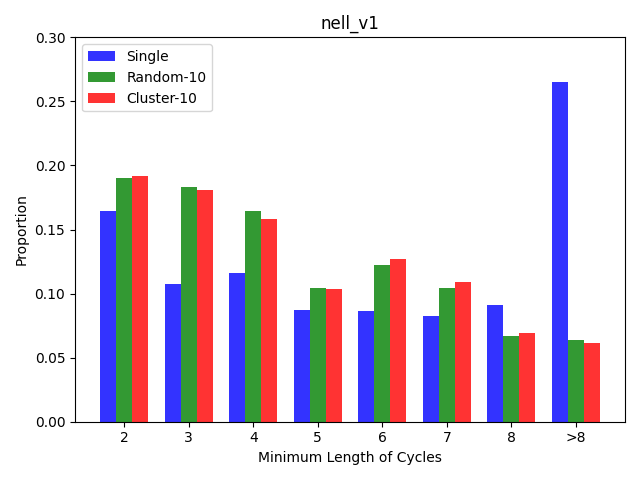}
		\end{minipage}%
	}%
	\centering
	\vspace{-.1in}
	\caption{Histogram of shortness on different datasets}
	\label{fig:shortness}
	\vspace{-.15in}
\end{figure}

\textbf{Evaluation of shortness.} Recall that in Section~\ref{subsec:pursuit}, we hypothesize that the desired cycle bases should generally contain short cycles. In this paragraph, we evaluate the shortness of the SPT cycle bases on various datasets and analyze the correlation between the shortness and performance of different choices of cycle bases. To be specific, we draw histograms to evaluate the minimum length of cycles that pass a triplet. We compare different choices of cycle bases, including a single cycle basis, 10 randomly chosen cycle bases, and 10 cycle bases chosen by the clustering algorithm, which are denoted by "Single", "Random-10", and "Cluster-10" respectively. The histograms are shown in Figure~\ref{fig:shortness}. In the histogram, the x-axis denotes the minimum length of cycles that pass a certain triplet, and the y-axis represents the proportion of triplets with a certain minimum length of cycles among all triplets.

As shown in Figure~\ref{fig:shortness}, the cycle bases selected by the clustering algorithm generally contain small cycles compared with the randomly selected cycle bases or the single cycle basis. We can find that in Table~\ref{tab:ablation}, CBGNN outperforms CBGNN-Random on most datasets. Another interesting observation is that in Figure~\ref{fig:shortness} (d), the randomly selected cycle bases perform comparably with the cycle bases generated using the clustering algorithm in terms of shortness on NELL-995 v1. Recall that in Table~\ref{tab:ablation}, the performance of CBGNN-Random slightly beat CBGNN on NELL-995 v1. The above observations show the correlation between the shortness of cycle bases and their performance. The correlation may result from the fact that if a triplet is near to a tree root, then the cycles in the corresponding cycle basis that pass the triplet should be generally short. Most of the triplets are close to at least one tree root among the clustered cycle bases, and thus are easy to learn.

\end{document}